\documentclass[11pt,a4paper,logo]{googledeepmind}

\setleftlogo[180pt]{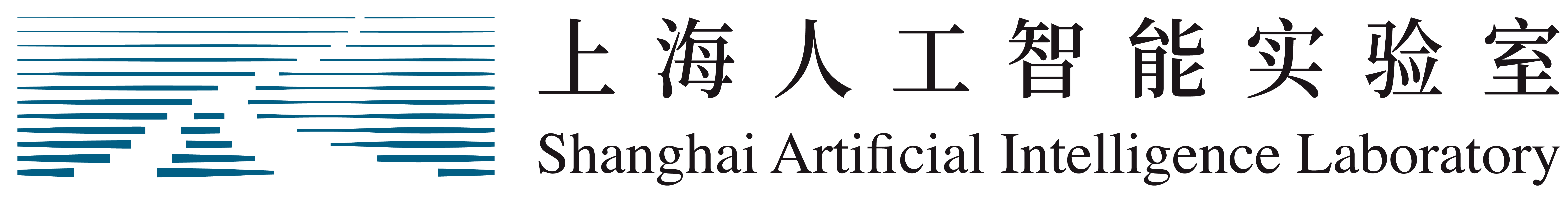} 
\setrightlogo[180pt]{}

\usepackage{placeins}

\usepackage[numbers,square,sort&compress]{natbib}

\usepackage{csquotes}

\newcommand{\ProjectName}{Sage-A1.5}
\newcommand{\tokenp}{\raisebox{-1.5pt}{\includegraphics[height=1em]{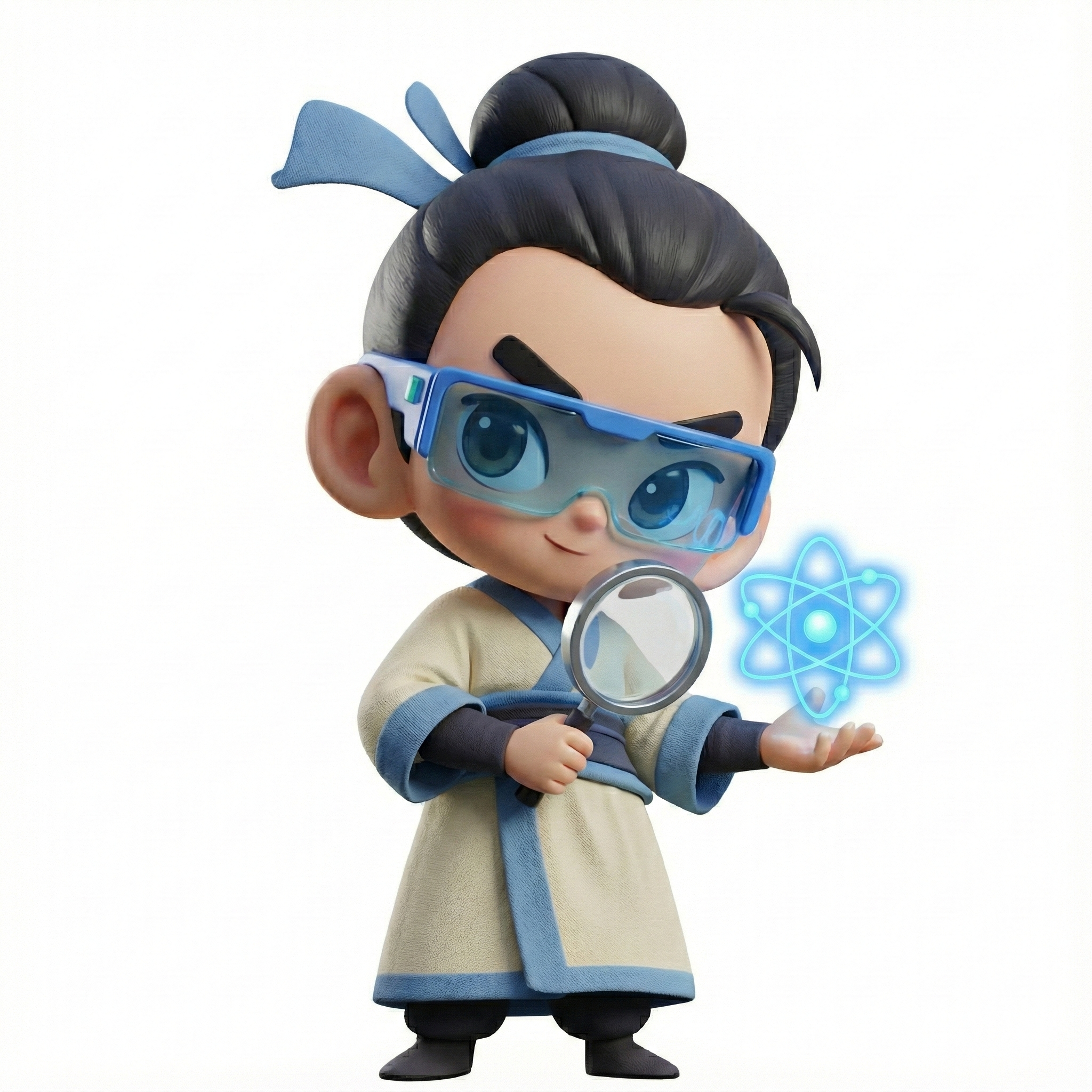}}}

\title{\ProjectName:~From Long-Horizon Execution to Scientific Exploration with Capability Composition}

\author[1]{\ProjectName~Team, Shanghai Artificial Intelligence Laboratory}

\usepackage{pdflscape}
\usepackage{amsthm}

\usepackage{textcomp}
\usepackage{rotating}
\usepackage{setspace}
\usepackage{soul} 
\usepackage{multicol}
\usepackage{subcaption}
\usepackage{caption}

\sethlcolor{green!14}

\usepackage{array}
\usepackage{multirow}
\usepackage{amsmath}
\usepackage{siunitx}

\usepackage{enumitem}
\usepackage{float}
\usepackage{seqsplit}
\usepackage{framed}
\usepackage{tikz}
\usepackage{listings}
\usepackage{colortbl}
\usepackage{wrapfig}
\usepackage[most]{tcolorbox}
\usepackage{pdfpages}
\tcbuselibrary{listingsutf8}
\definecolor{mycolor}{RGB}{50,80,150}

\usepackage{ragged2e}
\usepackage{makecell}
\usepackage{adjustbox}

\usepackage{fontawesome5}
\definecolor{homepageblue}{HTML}{247BA0}

\usepackage[symbol]{footmisc}
\newcolumntype{Y}{>{\RaggedRight\arraybackslash}X}

\hypersetup{
    colorlinks=true,
    linkcolor=blue, 
    citecolor=blue,  
    filecolor=black,
    urlcolor=blue    
}
\definecolor{AbstractBgColor}{HTML}{F4F7FB}
\usepackage{tocloft}

\usepackage{etoolbox}
\usepackage{arydshln}
\makeatletter
\patchcmd{\@tocline}
    {\hfil}
    {\leaders\hbox{\hfil}\hfil}
    {}{}
\makeatother

\begin{document}
\sloppy
\thispagestyle{firststyle}

\begin{tcolorbox}[
    colback=AbstractBgColor, 
    colframe=AbstractBgColor, 
    arc=5pt,                  
    auto outer arc,
    boxrule=0pt,              
    left=8pt, right=8pt, 
    top=4pt, bottom=4pt,      
    parbox=false,
    width=\textwidth,
    before skip=-800pt,        
    after skip=0pt, 
    enlarge top by=-12pt
]
\begin{abstract}
\vspace{-14pt}
In this report, we introduce \ProjectName, a 35B Mixture-of-Experts agentic model that builds on A1 to advance long-horizon problem solving through adaptive capability composition. Compared with A1, \ProjectName~learns not only to solve individual tasks, but also to organize work and revise its approach during execution. \ProjectName~highlights three improvements: agentic orchestration, verification, and open scientific discovery. First, agentic orchestration extends single-agent execution through task delegation and parallel collaboration among subagents with isolated contexts, enabling complex tasks to be decomposed and addressed concurrently and supporting efficient execution while reducing interference across concurrent subagents. Second, agentic verification strengthens long-horizon reliability through progress checking, evidence validation, and reflective strategy revision, allowing the agent to recognize unproductive search paths and pursue alternative approaches. Third, open scientific discovery extends agentic capabilities beyond answering predefined questions to conducting exploratory research, connecting scientific inquiry with interactive analysis and inspectable research artifacts. We showcase these capabilities through complicated agentic tasks and multidisciplinary scientific applications. Besides, we hope \ProjectName~provides the community with an accessible foundation for open scientific discovery, offering a practical path from long-horizon task execution to collaborative, verifiable, and inspectable scientific exploration.

\end{abstract}

\newpage

\maketitle

\begin{center}
\href{https://alpha-innovator.github.io/Sage-A1.5}{\raisebox{-0.15\height}{\includegraphics[height=1em]{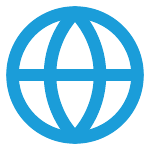}}~\textcolor{black}{\textbf{Homepage}}} \qquad
\href{https://discovery.intern-ai.org.cn/token-plan/home?tabIndex=0}{\textcolor{black}{\tokenp~\textbf{Token-plan}}} \qquad
    \href{https://huggingface.co/InternScience/Sage-A1.5}{\raisebox{-0.15\height}{\includegraphics[height=1em]{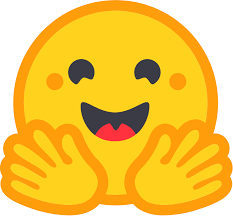}}~\textcolor{black}{\textbf{Model}}}
\end{center}
\end{tcolorbox}

\begin{figure}[H]
    \centering
    \includegraphics[width=\linewidth]{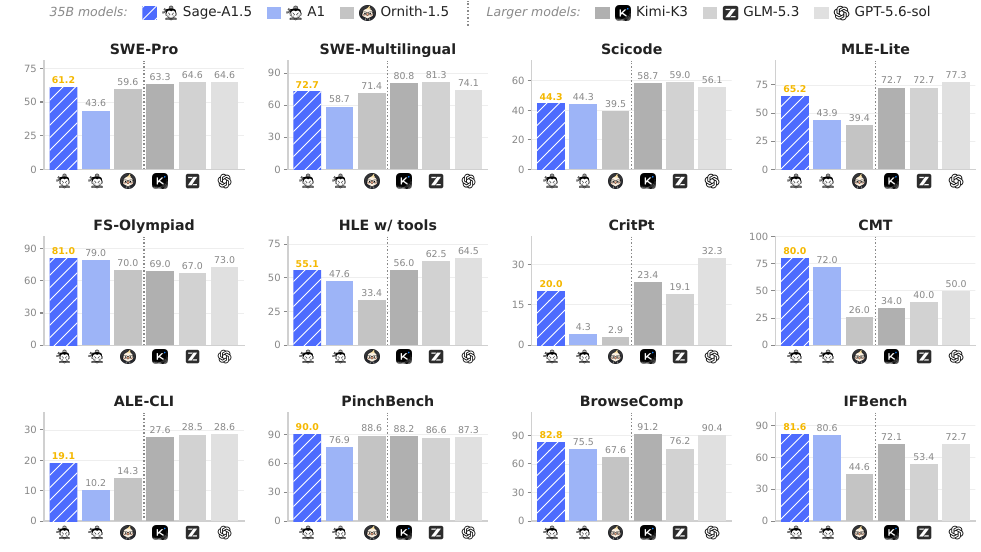}
    \caption{Benchmark performance of \ProjectName.
    }
    \label{fig:benchmark_performance}
\end{figure}

\tableofcontents

\newpage

\section{Introduction}
\label{sec:introd}

In our earlier work, we introduced Agents-A1~\citep{bai2026scaling}, a 35B Mixture-of-Experts (MoE) agent model that improves long-horizon task performance by scaling the agent horizon rather than model parameters. We built a knowledge–action infrastructure~\citep{cao2026agents} that connects external knowledge, tool calls, observations, and verification feedback, producing training trajectories with an average length of 45K tokens. Agents-A1~\citep{bai2026scaling} showed strong performance in multi-step search, scientific reasoning, instruction following, and tool use, matching or exceeding trillion-parameter models on several benchmarks. These results established a practical path toward capable long-horizon agents at a smaller model scale.

Recent progress~\citep{GLM-5.3, kimik3, GPT-5.6, cao2026agents} demonstrates that strong domain skills alone are not enough for complicated tasks. Software engineering, machine learning optimization, and scientific research often require agentic model to manage related subtasks, compare different approaches, and revise decisions as new evidence becomes available. A single agentic model working step by step may struggle with a growing context, slow exploration, and errors that affect later decisions. Longer execution also does not always lead to better results: an agent may repeat unproductive searches, build on unsupported assumptions, or accept a conclusion before checking all the evidence. Further progress therefore requires not only stronger intrinsic agentic capabilities and broader skills, but also better ways to organize agent work and check progress. \textit{Task delegation, parallel collaboration, evidence review, and strategy revision should become learned model capabilities rather than depend only on fixed rules in an external framework.}

To address these challenges, we introduce \ProjectName{}, a 35B MoE agent model built on Agents-A1, with two key capabilities: \textbf{Agentic Orchestration} and \textbf{Agentic Verification}. \textbf{1)} Agentic Orchestration allows the model to decide how to divide work, delegate subtasks to parallel subagents with separate contexts, and combine their findings toward a shared goal. \textbf{2)} Agentic Verification allows the model to recognize when progress has stalled or an important conclusion needs checking, review the available evidence and assumptions, and adjust its next actions. At inference time, this verification takes place within the model’s own reasoning and tool-use process, without an online supervisor or a separate verification tool. These capabilities offer the community a scalable technical pathway for translating model capabilities into agentic capabilities. As shown in Fig.~\ref{fig:benchmark_performance}, despite having only 35B parameters, \ProjectName~outperforms the substantially larger Kimi-K3~\citep{kimik3}, GLM-5.3~\citep{glm5}, and GPT-5.6-sol~\citep{GPT-5.6} on FrontierScience-Olympiad~\citep{wang2026frontierscience}, CMT~\citep{pan2026cmtbenchmarkbenchmarkcondensedmatter}, PinchBench~\citep{pinchbench2026}, and IFBench~\citep{pyatkin2025ifbench}, while remaining competitive with them on SWE-Bench Pro~\citep{deng2025swebenchpro} and BrowseComp~\citep{browsecomp}.
Moreover, \ProjectName~matches or surpasses the equally sized Agents-A1~\citep{bai2026scaling} and Ornith-1.5~\citep{ornith_1_5} on all twelve benchmarks.

Our contributions include the following three aspects:
\begin{itemize}
    \item We develop Adaptive Agentic Execution, integrating \textbf{Agentic Orchestration} and \textbf{Agentic Verification} into \ProjectName’s execution policy for autonomous coordination of parallel subtasks and evidence-driven self-correction throughout long-horizon problem solving.
    \item We build a unified training pipeline that combines multi-domain supervised fine-tuning with mixed-reward reinforcement learning, fusing capabilities across domains, reasoning modes, and agent harnesses into a single agentic model.
    \item We demonstrate \ProjectName{}'s potential for agentic scientific discovery through broad benchmark evaluations and scientific case studies, highlighting a shift from isolated task solving toward long-horizon scientific research.
\end{itemize}
 
\section{Method}

\begin{figure}[t]
    \centering
    \includegraphics[width=\linewidth]{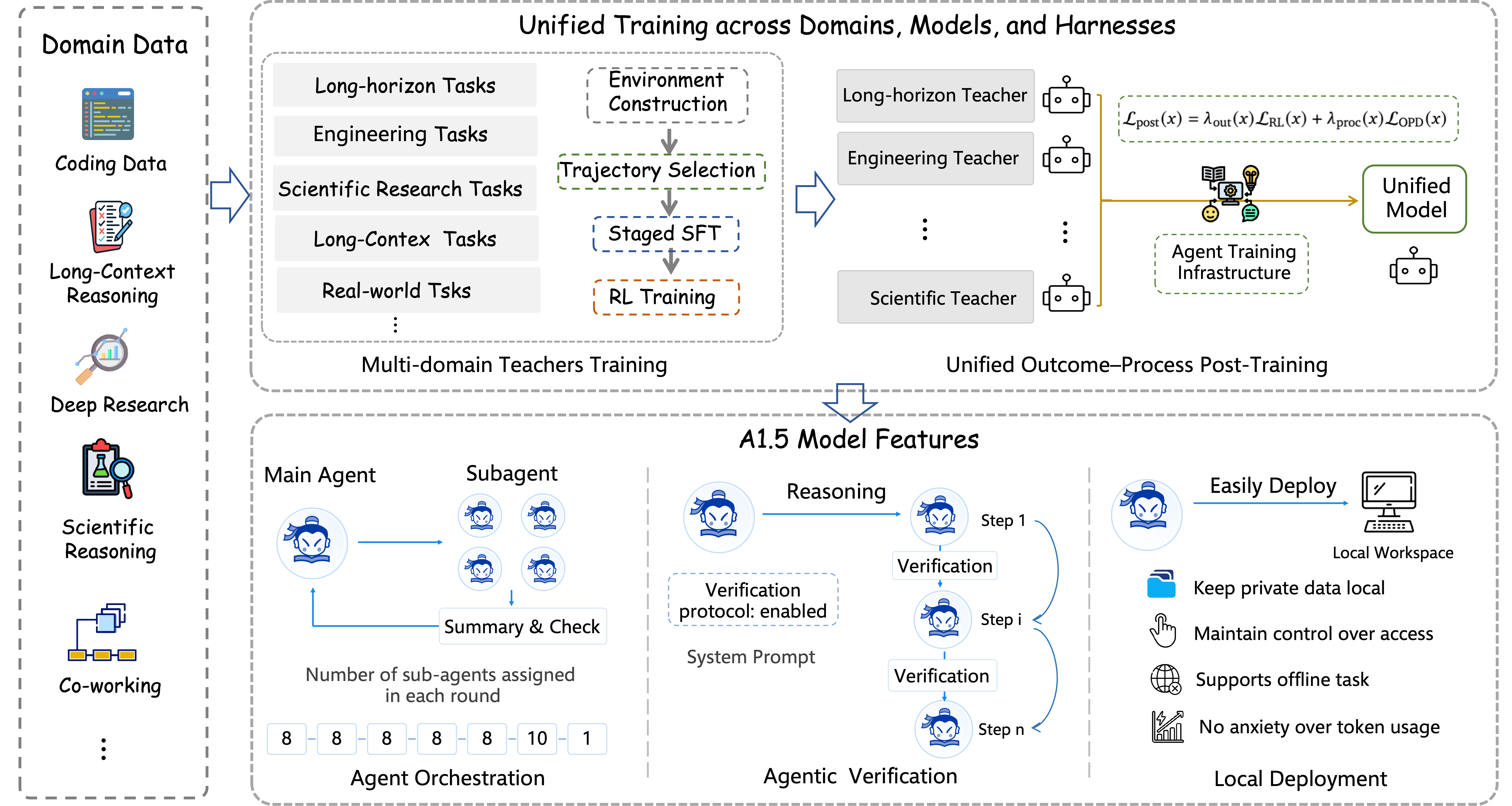}
    \caption{Overview of the three-stage training pipeline of \ProjectName. First, the model is trained on diverse domain data, including coding, long-context reasoning, deep research, scientific reasoning, and co-working, utilizing environment construction, trajectory selection, staged supervised fine-tuning, and reinforcement learning to build multi-domain teachers. Then, these teachers undergo unified outcome-process post-training to produce the final unified model, which features an agent orchestration for collaborative task solving, iterative agentic verification for reasoning reliability, and local deployment for ensuring data privacy, access control, and offline support.
    }
    \label{fig:training_framework}
\end{figure}

\subsection{Adaptive Agentic Execution: From Capability Coverage to Capability Composition} 
\label{sec:capa_compos}
\begin{figure}[htbp]
    \centering
    \includegraphics[width=\linewidth]{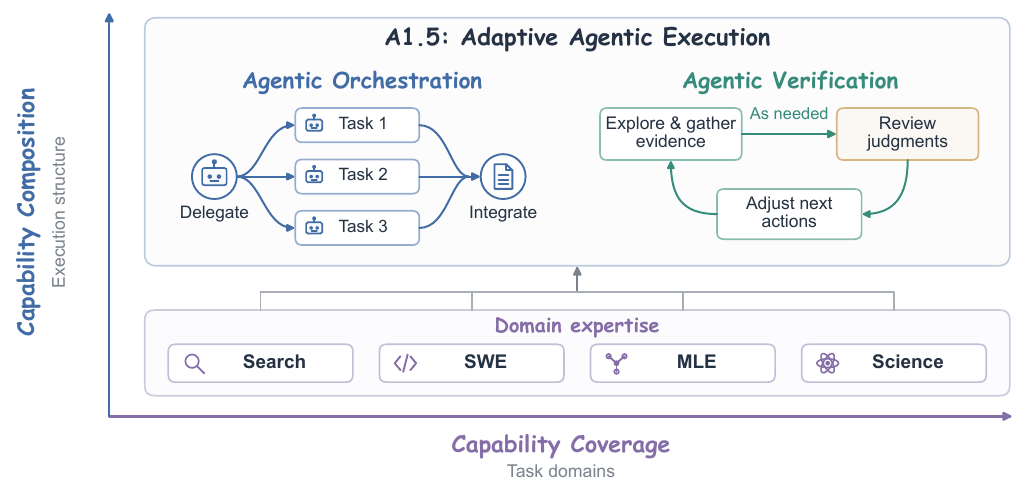}
    \vspace{-15pt}
    \caption{Two complementary dimensions of agent capability. Capability coverage spans task domains, while capability composition concerns how problem-solving capabilities are organized and adapted during execution. Agentic Orchestration coordinates parallel work and integrates results; Agentic Verification reassesses intermediate responses using available evidence and adjusts subsequent actions.}
    \label{fig:capability-dimensions}
\vspace{-10pt}
\end{figure}

Domain-specific training expands capability coverage---the range of domains and tasks an agent can address. In \ProjectName{}, we extend agent capability development along a second dimension, \textit{\textbf{capability composition}}: how problem-solving capabilities are organized, coordinated, and adapted during execution. This dimension concerns execution structure across task domains. We develop it through Agentic Orchestration and Agentic Verification (Fig.~\ref{fig:capability-dimensions}). Agentic Orchestration determines how work is decomposed, delegated, and integrated across agent executions. Agentic Verification determines when to reassess intermediate judgments and exploration strategies, using the findings to revise subsequent actions. As a result, these capabilities make execution organization and feedback-driven adjustment explicit targets of model training alongside domain-specific problem solving.

\subsubsection{Agentic Orchestration} 
\label{sec:agentic-orchestration}

We propose \textbf{Agentic Orchestration}, a foundational capability of \ProjectName{} for structuring, allocating, and coordinating computation during task execution. The key insight is to make how execution is organized an explicit target of model training: \textit{learning what work to pursue and how to combine the results complements learning to solve individual tasks.} 

\noindent \textbf{Capability Definition and Composition Patterns.}
Agentic Orchestration is the model's ability to organize multiple agent executions around a shared objective. Its policy determines whether and how to decompose work, which subtasks to delegate, what context to provide, and how to coordinate dependencies and integrate results. The model adapts subsequent work to returned evidence, connecting local execution to the overall task. We distinguish two complementary patterns of parallel agent execution:

\begin{itemize}[leftmargin=1.5em,itemindent=0pt,listparindent=0pt,itemsep=0.4em,topsep=0.4em,parsep=0pt]
    \item \textbf{Exploratory Composition.} The model assigns multiple agents different hypotheses or solution approaches to the same question. Their outputs provide alternative candidates and supporting evidence, which the main agent compares to narrow the investigation and decide which directions to pursue. For example, agents may explore different system architectures against the same requirements, returning candidate designs for comparison.

    \item \textbf{Constructive Composition.} The model decomposes an overall task into complementary subproblems and assigns them to agents for parallel work. Each output contributes a distinct part of the solution; the main agent integrates these parts, coordinating dependencies and checking coverage and consistency. For example, agents may develop the data, interface, and deployment components of one system design, which must fit together to meet the overall requirements.
\end{itemize}

Exploratory and constructive composition are analogous to \(A \lor B \lor C\) and \(A \land B \land C\), respectively. Here, \(A\), \(B\), and \(C\) denote tasks delegated to agents: alternative investigations of a shared question in the former, and complementary subproblems in the latter. Complex tasks dynamically interleave both patterns as emerging evidence informs task decomposition, exploration, and integration.

We develop the Agentic Orchestration capability through two training task families: \textit{Constraint-Guided Long-Horizon Search} and \textit{Multi-Source Research Synthesis}. \textbf{1)} Long-horizon search tests whether the model can choose productive angles for parallel investigation, delegate independent investigations of a shared target's relation constraints, and use returned evidence to narrow candidate answers and guide subsequent searches. \textbf{2)} Research synthesis tests whether it can divide a broad research objective into complementary parallel assignments, assess the coverage and consistency of their findings, and integrate them into a report that answers the original research question.

\begin{figure}[htbp]
\vspace{-8pt}
    \centering
    \includegraphics[width=\linewidth]{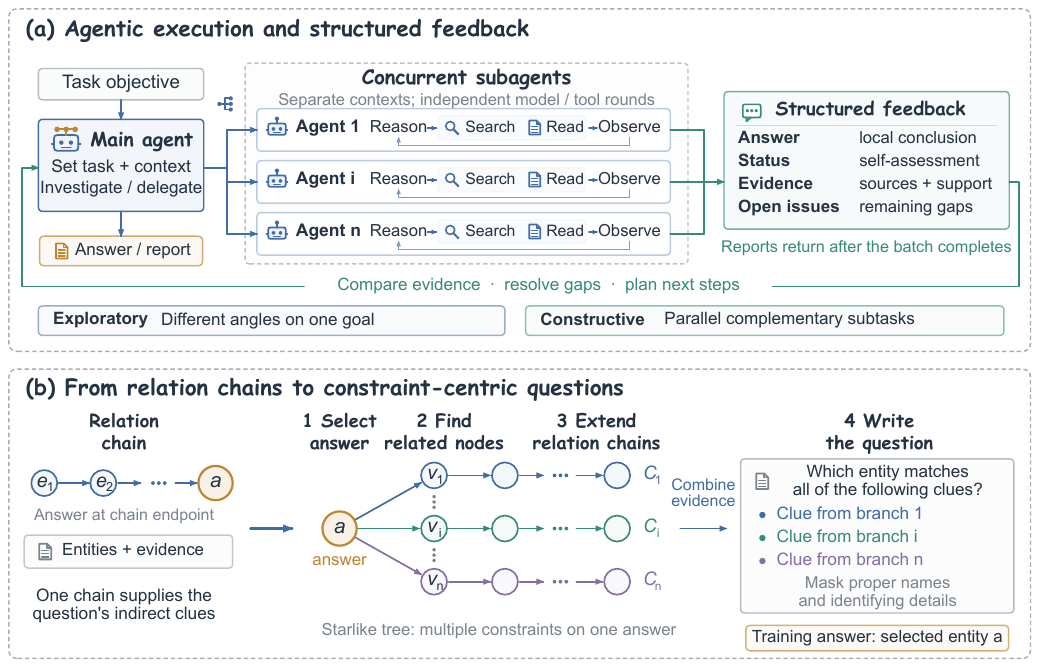}
    \vspace{-14pt}
    \caption{Agentic Orchestration in execution and task construction. (a) Independently executing subagents return structured feedback for the main agent's next decisions. (b) Multiple relation chains are built around a selected answer and combined into one question, with identifying details masked. Their clues provide complementary constraints for parallel investigation.}
    \label{fig:agentic-orchestration}
\vspace{-10pt}
\end{figure}

\noindent \textbf{Orchestration Protocol and Structured Feedback.}
The main agent investigates directly or delegates subtasks, specifying their scope and context. Subagents search and read webpages in separate contexts, advancing their model calls and tool interactions independently within a concurrent batch. Once the batch completes, the main agent uses the findings to synthesize an answer or organize further investigation (Fig.~\ref{fig:agentic-orchestration}a).

\begin{figure}[htbp]
\vspace{-6pt}
    \centering
    \includegraphics[width=\linewidth]{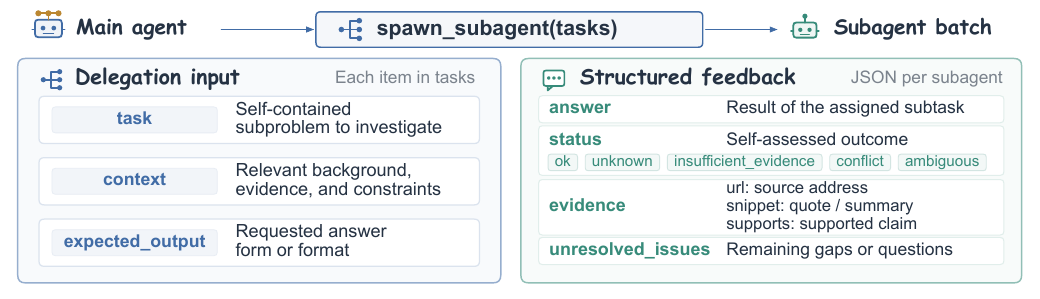}
    \vspace{-12pt}
    \caption{Delegation interface and structured subagent feedback.}
    \label{fig:orchestration-protocol}
\vspace{-8pt}
\end{figure}

Building on the model's ability to critique its own execution, we ask each subagent to return its result together with a self-assessment, supporting evidence, and remaining issues. This feedback helps the main agent identify gaps and conflicts and provides signals for trajectory selection. Fig.~\ref{fig:orchestration-protocol} summarizes the interface and structured feedback format.

\noindent \textbf{Task Construction: From Relation Chains to Constraint-Centric Graphs.}
For Constraint-Guided Long-Horizon Search, we extend the relation-chain construction method in Agents-A1~\cite{bai2026scaling}. We select an answer entity, identify directly related entities that provide different relation constraints, and extend a chain from each of them. This forms a \textbf{starlike tree graph} around the shared answer (Fig.~\ref{fig:agentic-orchestration}b). Branch descriptions and evidence are combined into one question, with proper names and identifying details masked. The main agent can delegate different constraints for parallel investigation, explore candidates from multiple angles, and combine the findings to recover the shared answer.

Multi-Source Research Synthesis combines evidence from multiple sources into a coherent research report. Agentic Orchestration supports complementary research assignments, comparison of findings across sources, and follow-up investigations to fill gaps or resolve inconsistencies before synthesis. Task construction is described in Section~\ref{sec:search-agents}.

\noindent \textbf{Trajectory Collection and Role-Specific Supervision.}
Teacher model runs within the orchestration harness with role-specific prompts, tools, and execution budgets. It chooses when to investigate directly, delegate, and follow up on returned evidence. We record the complete executions and treat the main-agent trajectory and each subagent trajectory as independent agent samples.

Main-agent samples teach task decomposition, delegation, use of feedback, and synthesis; subagent samples teach independent investigation, evidence organization, and structured reporting. This supervision complements domain expertise with the ability to structure, allocate, and coordinate computation during agentic execution.

\subsubsection{Agentic Verification} 
In long-horizon research tasks, agents must gradually build chains of evidence through multiple rounds of retrieval and reasoning. When an exploration path repeatedly fails to yield useful information, a model may continue searching around the same hypothesis or use unverified assumptions as premises for subsequent reasoning, leading to ineffective exploration and accumulating errors. To address this problem, we propose \textbf{Agentic Verification}, which enables the model to adaptively recognize when verification is needed, examine existing evidence and exploration strategies within its own reasoning, and adjust subsequent actions accordingly. When progress stalls or an important intermediate conclusion is reached, the model reviews facts confirmed by evidence, examines earlier judgments for unsupported inferences, identifies unresolved information gaps, and determines which exploration paths have proved ineffective and how to proceed. The central idea is to make verification a learned capability of the model itself: \textit{the model adaptively determines when to verify, what to verify, and how to continue based on the findings, rather than having verification triggered and executed by an external harness according to predefined rules.} This mechanism enables the model to proactively revise its judgments, gather additional evidence, or change its search direction, improving its ability to self-correct during long-horizon tasks.

To train this capability, we construct complete research trajectories containing intermediate reflections and the actions that follow them. During data construction, an execution model conducts research through live search and webpage reading, while a supervisor model examines the exploration process and introduces verification at points involving repetitive searches, sustained lack of progress, or key conclusions requiring confirmation. Each verification relies only on the evidence available at that point, distinguishes confirmed facts from unverified hypotheses, and proposes a concrete direction for further investigation. The execution model then continues exploring based on the verification findings, producing coherent training examples that connect problem recognition, evidence assessment, and action adjustment. After data cleaning and format normalization, external supervision prompts and control messages are removed, and the reflections are incorporated directly into the model’s reasoning text. This allows the model to learn when to verify, what to examine, and how verification should inform subsequent actions. External supervision is used only to construct training data; at inference time, the model performs this process independently. 

\textbf{Note that} we also introduce a conditioning marker during training to enable the capability on demand: for models trained in this way, adding \texttt{Verification protocol:enabled} to the system prompt activates Agentic Verification. Once enabled, the model autonomously performs verification within its regular reasoning and tool-use process, without requiring an additional verification tool, an online supervisor model, or a verification procedure implemented by the harness.

Building on this approach, we plan to integrate verification with context management, using intermediate verification points as natural boundaries for organizing the exploration history. After completing an intermediate reflection, the model will autonomously determine whether the preceding exploration segment can be consolidated into a concise stage record. If so, the corresponding history segment will be replaced with the stage record and the current verification, while preserving key evidence, rejected hypotheses, unresolved questions, and the current task state. Compression will apply only to the segment between the two most recent verification points, avoiding repeated compression of earlier history. Through this design, we aim to control context overhead while preserving reasoning continuity and evidence integrity, supporting research tasks over longer horizons.

\subsection{Post-training on Engineering Tasks}
\subsubsection{MLE Tasks}
\noindent\textbf{Task Construction.}
We design Kaggle-MLE Factory, an automated multi-agent pipeline that converts historical Kaggle competitions into executable tasks with public workspaces, hidden test sets, and a unified evaluation interface. The pipeline assigns distinct roles to competition selection and data acquisition, task analysis, code generation, and execution and validation. The analysis module combines official descriptions with the actual data to form task specifications. The generation module then prepares the data, constructs held-out test splits with hidden labels, and implements task-specific evaluators. The execution and validation modules test the generated tasks on real data and provide feedback for iterative repair. Tasks undergo human review before trajectory collection. We also use MLE-Dojo~\cite{mledojo} as a supplementary task source and exclude competitions that overlap with the full MLE-Bench task set~\cite{mlebench}.

\noindent\textbf{Interaction Harness.}
We build an MLE interaction harness for long-horizon experimentation, allowing the model to autonomously refine solutions based on execution feedback. The harness organizes candidate code and execution results into an experiment tree that records solution revisions, supporting branch comparison, backtracking, submission management, and invalidation of candidates with unreliable results. Persistent plans and context compression support sustained interaction, while an analysis sub-agent with an independent context provides data exploration and result diagnosis.

\noindent\textbf{Trajectory Collection and Selection.}
We select demonstration trajectories and useful supervision segments based on the actual execution and evaluation results of candidate solutions. If performance degradation after the first best-performing submission exceeds a threshold, we truncate the trajectory at a complete tool-interaction boundary near that submission; otherwise, we retain it through the final submission. This preserves useful exploration and failure recovery while removing persistently degrading tails. Supervision data include initial-solution prefixes that execute successfully and meet quality criteria, long-horizon optimization trajectories, and analysis sub-agent trajectories, covering solution construction, iterative refinement, and experimental diagnosis. Training retains prior interaction context and supervises only newly generated reasoning, text, and tool calls.

\noindent\textbf{Staged Supervised Fine-Tuning.}
We adopt two-stage supervised fine-tuning to adapt general interaction capabilities to MLE. The first stage uses general coding and tool-interaction trajectories to establish code revision and tool-use capabilities. The second stage combines MLE trajectories with experimental trajectories from literature-derived scientific tasks to train solution design, result analysis, and sustained optimization, initializing the policy for online reinforcement learning.

\noindent\textbf{Windowed Interaction Reinforcement Learning.}
Starting from SFT, we apply Windowed Interaction Reinforcement Learning (WIRL). Since each experimental outcome reflects multiple rounds of analysis, code revision, and tool interaction, WIRL groups interactions between successive experiment results into execution windows with shared outcome-based rewards. Decisions remain conditioned on accumulated history, while completed windows support policy updates before the task ends.

Window rewards combine solution quality measured by task-specific medal thresholds with performance changes relative to the most recent valid submission. Model-induced execution errors and unscorable submissions receive negative rewards. Each window reward directly weights its main-model output tokens during policy optimization, without group-relative normalization.

\noindent\textbf{Asynchronous Policy Optimization.}
We asynchronously collect completed windows with trusted feedback and correct policy differences using token probability ratios between the pre-update and rollout policies. Ratios within a specified interval serve as correction weights; tokens outside the interval receive zero policy-gradient weight. For a window batch $\mathcal B$, we optimize
\begin{equation}
\label{eq:mle-wirl-objective}
\mathcal L(\phi)=
\frac{1}{|\mathcal B|}
\sum_{u\in\mathcal B}
\frac{1}{\max(N_u,1)}
\sum_t m_t
\left[
w_t\,\ell_t(\phi;R_u)
+\lambda_{\mathrm{KL}}K_t
-\lambda_{\mathrm H}H_t
\right].
\end{equation}
Here, $\ell_t(\phi;R_u)$ is the PPO clipped loss~\cite{ppo} with window reward $R_u$ as its advantage input, and $w_t$ is the stop-gradient correction weight. The terms $K_t$ and $H_t$ denote KL regularization against a fixed SFT policy and current-policy entropy, weighted by $\lambda_{\mathrm{KL}}$ and $\lambda_{\mathrm H}$. The mask $m_t$ selects newly generated main-model outputs, with $N_u=\sum_t m_t$. The loss is normalized by valid output tokens within each window, then averaged across windows.

\subsubsection{SWE Tasks}
We construct training data for repository-level software-engineering issue resolution. Each task provides a repository snapshot and a natural-language issue description, and the agent interacts with the repository to produce a patch. The patch is evaluated against held-out tests: designated failing tests must pass, while previously passing tests must remain passing.

\textbf{Task Sources and Environment Construction.}
The task pool consists of instances from the public SWE-rebench-V2 dataset~\citep{badertdinov2026swerebenchv2languageagnosticswe} and a separate set of tasks constructed in house. We convert tasks from both sources into a common repair-task format following SWE-bench~\citep{deng2025swebenchpro}. Each task starts from the base commit of its associated pull request and uses the corresponding issue description as the task prompt. The reference diff is split into a solution patch containing changes to non-test files and a test patch reserved for grading.
For each task, we recover or construct the container configuration, installation recipe, and test command. Compatible tasks within a repository release group share environment configurations, with dependency mirrors and runtime settings adapted to our cluster. When a recipe is missing or unusable, we follow the build-and-repair loop of SWE-Factory~\citep{guo2025swefactory}. Starting from a validated configuration for a neighboring version when available, we build the container, run the tests, and revise the configuration based on the resulting error logs for a bounded number of rounds. Reference-patch size and modified-file count are used to prioritize task collection rather than as hard exclusion criteria.

\textbf{Task Validation.}
We run both reference-patch and no-op checks on candidate tasks. The reference-patch run checks whether the known fix satisfies the grading criteria in the prepared environment. The no-op run submits an empty patch and checks that it receives zero reward, flagging tasks that can receive reward without a repair. We record both outcomes as task metadata for curation.

\textbf{Trajectory Collection.}~We collect trajectories using two agent harnesses: \texttt{mini-swe-agent}~\citep{swe-agent}, which exposes a single shell-command action per turn, and Claude Code, which provides dedicated editing and shell tools. When both harnesses solve the same task, we retain both trajectories. Rollouts run in per-task isolated containers with turn and wall-clock limits.

Each zero-reward trajectory is re-graded once by replaying its unchanged patch, without further agent interaction, to recover successful patches affected by transient execution or grading failures. If the replay passes, we update the trajectory's reward label.

The training corpus includes successful trajectories and a filtered subset of unresolved attempts. For unresolved attempts, we require a non-empty patch, substantive task-directed interaction, well-formed structure, a passing leakage check, and additional task or harness coverage. Runs consisting solely of infrastructure failures or bare timeouts are excluded.

\textbf{Leakage Control.} We restrict access to reference fixes and held-out tests during agent execution. Before each rollout, we remove Git objects that expose these artifacts; resetting the working tree to the base commit alone does not remove them from the object store. Because the grader retrieves the held-out tests from the reference commit, the required objects are restored after the agent exits. The held-out tests are introduced only at grading time.

All outbound traffic from rollout containers is forced through the cluster proxy. During agent execution, we redirect the proxy endpoint to a closed local port to block external access. The separate grading stage retains the connectivity needed to install dependencies. These controls apply to data collection, evaluation, and training-time environment interaction.

We audit trajectories for evidence that reference-solution content reached the agent. Network commands alone are not classified as leakage, since they may be used for dependency installation or connectivity checks. Audit verdicts are stored with the trajectories and used during selection. Instances overlapping the test splits of our evaluation suites are removed before collection.

\subsection{Scientific Research}

\begin{figure}
\vspace{-10pt}
    \centering
    \includegraphics[width=1\linewidth]{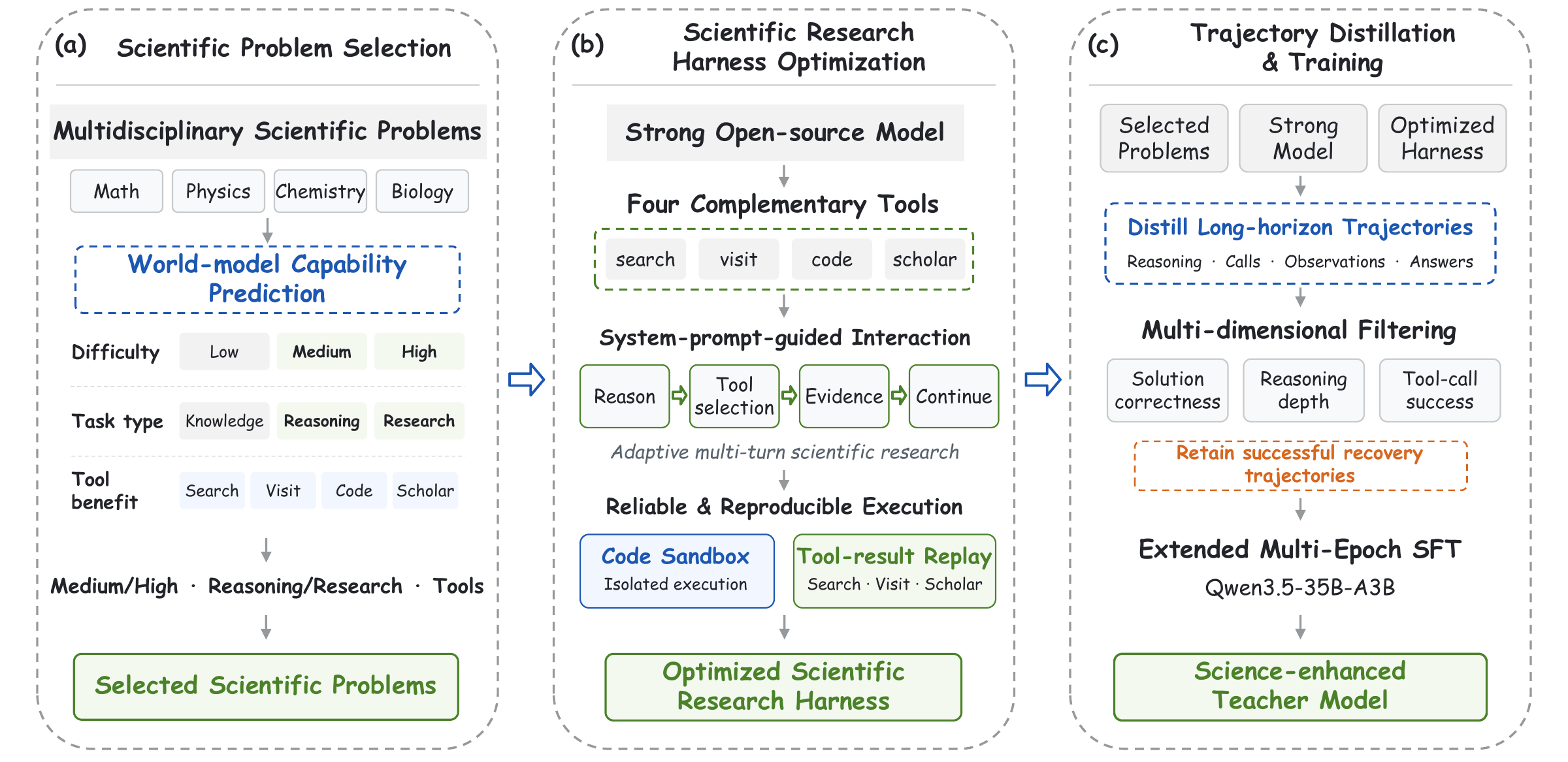}
    \vspace{-15pt}
    \caption{Overall framework of the three-stage science-enhanced teacher model training pipeline, comprising capability-guided scientific problem selection, scientific research harness optimization, and trajectory distillation and training to enhance long-horizon scientific research capabilities.}
    \label{fig:method_scientific_research}
\vspace{-10pt}
\end{figure}

Scientific research requires models to combine deep scientific reasoning, effective tool use, and long-horizon problem-solving capabilities. To enhance these capabilities, we develop a science-enhanced teacher model through a three-stage pipeline~\cite{ye2026exomind}, as illustrated in Fig.~\ref{fig:method_scientific_research}. First, \emph{scientific problem selection} identifies high-quality scientific problems based on their difficulty, task type, and expected benefits from tool use. Second, \emph{scientific research harness optimization} improves the tool-use framework to enable reliable, efficient, and reproducible interactions with external environments. Third, \emph{trajectory distillation and training} generates long-horizon reasoning and tool-use trajectories using a strong model with the optimized harness, selects high-quality demonstrations through multi-dimensional filtering, and performs tool-augmented supervised fine-tuning. Together, these stages establish a unified pipeline that integrates high-quality problem selection, robust tool interaction, and effective model training to strengthen scientific research capabilities.

\noindent \textbf{Scientific Problem Selection.}
We first collect a diverse pool of scientific problems spanning mathematics, physics, chemistry, biology, and related disciplines, followed by preliminary quality checks to remove invalid or inconsistent instances. Inspired by ExoMind~\cite{ye2026exomind}, we leverage the world-model predictive capabilities of a strong model to evaluate each candidate problem along three dimensions: \textbf{(i) Difficulty}, classified as low, medium, or high; \textbf{(ii) Task type}, categorized as knowledge-oriented, reasoning-oriented, or research-oriented; and \textbf{(iii) Tool benefit}, estimating the potential benefits of external tools, including \texttt{search}, \texttt{visit}, \texttt{code}, and \texttt{scholar}, for problem solving. Based on these assessments, we prioritize medium-to-high difficulty problems that require substantial reasoning or research-oriented exploration and are likely to benefit from tool interactions. This capability-guided selection yields a high-quality scientific training corpus characterized by problem validity, strong relevance to scientific reasoning and research, and close alignment with the target model's reasoning and tool-use capabilities.

\noindent\textbf{Scientific Research Harness Optimization.} 
Building on a strong open-source model, we optimize a scientific research harness equipped with four complementary tools: \texttt{search} for web retrieval, \texttt{visit} for accessing and inspecting web pages and documents, \texttt{code} for computational analysis and executable verification, and \texttt{scholar} for academic literature retrieval. We enhance the system prompt to guide scientific reasoning, adaptive tool selection, evidence integration, and multi-step interactions. To improve execution reliability and security, we introduce an isolated code execution sandbox that restricts access to host-local resources and provides a consistent execution environment, mitigating information leakage risks and environment-related failures. Furthermore, we implement a tool-result replay mechanism for \texttt{search}, \texttt{visit} and \texttt{scholar}, which reuses recorded observations for repeated or semantically equivalent requests under consistent retrieval contexts. This mechanism reduces redundant tool calls, improves the stability and reproducibility of tool interactions, and enhances the efficiency of long-horizon scientific research.

\noindent \textbf{Trajectory Distillation and Training.}
To develop the science-enhanced teacher model, we adopt a training pipeline comprising two key components: multi-dimensional trajectory filtering and extended multi-epoch training. First, we leverage a strong model with the optimized scientific research harness to distill long-horizon trajectories on the selected high-quality, challenging scientific problems, capturing intermediate reasoning, tool calls, observations, and final answers. We then filter these trajectories along three dimensions: \emph{solution correctness}, \emph{reasoning depth}, and \emph{tool-call success rate}, prioritizing demonstrations that combine accurate solutions, substantive reasoning, and effective tool use. Notably, we also retain selected trajectories involving recoverable tool failures, where the model identifies errors, adjusts its strategy, and successfully retries, providing valuable supervision for robust tool use and self-correction. Finally, we perform token-level supervised fine-tuning (SFT) on Qwen3.5-35B-A3B using the filtered trajectories. Given the high complexity, deep reasoning requirements, and extensive tool interactions of scientific research tasks, we adopt an extended multi-epoch training schedule to repeatedly expose the model to high-quality demonstrations. This sustained training aims to consolidate complex reasoning and tool-use behaviors, improving the stability of long-horizon scientific interactions while preserving the model's intrinsic reasoning capabilities.

\subsection{Recursive Task Evolution for Long-Context Reasoning}
\label{sec:recursive-task-evolution}

Long-context reasoning requires models to locate distributed evidence and integrate it through cross-passage relations and complex dependencies. However, generating questions directly from long documents can produce tasks that rely on local retrieval or shallow evidence aggregation. Repeatedly rewriting such questions may leave their underlying evidence dependencies and solution structure unchanged, yielding little increase in difficulty. We therefore adopt recursive task evolution driven by compositions of atomic capabilities. Diverse scenarios define the task objectives, while atomic capabilities characterize the operations required to solve them. Reference-model feedback then exposes shortcuts that guide subsequent revisions, directing difficulty growth toward changes in evidence organization and reasoning structure.

\noindent \textbf{Task Construction Across Diverse Scenarios.}
We construct questions from LongBlocks~\citep{ramos2026recipelongcontextreasoninglarge} documents spanning 16K--128K tokens, using 24 fine-grained task categories. These cover cross-document factual integration, timeline reconstruction, evidence attribution, multi-source consistency checking, rule induction, long-range state tracking, and version-impact analysis, among other scenarios. Each category specifies a solution objective, evidence scope, and answer format. For example, timeline reconstruction requires recovering event order, whereas version-impact analysis requires identifying affected components and their dependencies. We match documents to applicable tasks and generate questions, reference answers, and supporting evidence. This scenario-based organization provides explicit targets for task diversity while preserving a stable objective throughout subsequent evolution.

\noindent \textbf{Atomic Capabilities and Compositional Enhancement.}
Solving these tasks requires combining atomic capabilities such as evidence localization, alignment across surface forms, condition evaluation, candidate discrimination, set aggregation, state updating, and dependency propagation. Version-impact analysis, for example, involves locating relevant information, resolving aliases and version correspondences, propagating changes through dependencies, and applying exception conditions. We operationalize difficulty enhancement through ten axes: search-space size, quantifier and answer scope, predicate composition, positional non-locality of evidence, near-miss distractor strength, distractor dispersion, lexical gap between questions and evidence, state-transition depth, computational depth, and the depth of the shortest valid derivation. Each task uses applicable axes and construction operators to increase the demands on individual operations and their composition, while keeping the source document and task objective fixed.

\noindent \textbf{Feedback-Guided Recursive Evolution.}
The recursive process uses answerability as a constraint and reference-model performance as a selection signal. We estimate relative difficulty from repeated responses generated by Qwen3.5-35B-A3B. For reviewed questions that remain relatively easy, the generator receives the parent question, reference answer, and response feedback, and is instructed to modify at least two task-relevant enhancement axes to produce a new question, answer, and supporting evidence. Each child question undergoes evidence alignment and two-stage LLM review using the full context, followed by difficulty assessment. The review checks task conformity, answerability, evidential support, and answer correctness and completeness. Candidates that pass review and have an empirical success rate of at most $0.75$ enter the hard-question pool, while valid but easier candidates may seed further evolution. This process yields 3,512 hard QA pairs for subsequent trajectory sampling. Qwen3.5-35B-A3B achieves a mean per-question success rate of 31.57\% on this pool.
\noindent \textbf{Trajectory Generation and Joint Selection.}
Among correct solutions, we favor trajectories with complete evidential support and minimal redundant reasoning, aiming to teach the model to avoid unproductive context retrieval and repeated derivations. For each selected QA task, we sample three reasoning trajectories from DeepSeek-V4-Pro. We first judge final answers against the reference answer and discard incorrect candidates. The remaining trajectories for the same question are presented jointly to an LLM judge together with the original question and context. The judge decomposes the task into atomic requirements and evaluates evidence coverage, retrieval quality, necessary verification, reasoning consistency, and redundancy. To mitigate presentation-order bias, we conduct a second assessment with the candidate order reversed. Only candidates satisfying the strict quality criteria in both assessments are eligible for selection. We aggregate their scores and select one trajectory per question, preferring a more concise trajectory when quality is comparable. After applying sequence-length constraints, we retain 1949 LongBlocks trajectories for supervised fine-tuning.


\subsection{General Agents on Search and Real-World Workflows}

\subsubsection{Search Agents}
\label{sec:search-agents}

Deep search tasks require agents to decompose an open-ended information need, iteratively retrieve and read web content, and synthesize scattered evidence into a well-supported conclusion. Existing open-source search datasets mostly consist of short-answer questions with a single verifiable answer, which can be solved by locating one or a few key facts and provide limited supervision for broad exploration and evidence-grounded synthesis. We therefore complement open-source questions with open-ended research questions constructed by a search-grounded generation pipeline, and pair each question with explicit evaluation criteria that guide both trajectory collection and filtering.

\noindent\textbf{Research Question Collection.}
Research questions come from two sources. First, we collect questions from open-source search and deep-research datasets, followed by deduplication and preliminary quality checks to remove ambiguous, outdated, or ill-posed instances. Second, we construct new questions with a research question generation pipeline. Each generation task is seeded with a topic and a keyword, which jointly control topical diversity and anchor the question to a concrete information need. A question proposal agent equipped with web search and page reading tools then explores the seed by issuing queries, reading retrieved pages, and following leads across sources. Rather than relying on the parametric knowledge of the generator, we construct candidate research questions and their criteria from the proposal agent's trajectories, so that each question is grounded in information that is actually discoverable on the web and the criteria reflect the evidence gathered during exploration. The resulting questions are open-ended, requiring the integration of multiple pieces of evidence rather than the retrieval of a single fact, and each criterion specifies a concrete aspect that a satisfactory answer should cover, such as key facts, supporting evidence, comparisons, or conclusions.

\noindent\textbf{Trajectory Generation.}
Using the candidate research questions and their criteria, we collect agentic search trajectories in the same tool environment. For each question, the agent is instructed to conduct multi-turn search and reading, and to produce a detailed report that answers the research question and presents the supporting evidence, including the sources from which key claims are derived. This report-style output encourages the agent to explore the question broadly, cross-check information across sources, and ground its conclusions in retrieved evidence rather than unsupported claims.

\noindent\textbf{Criteria-Based Trajectory Filtering.}
An LLM evaluator assesses each trajectory against the criteria associated with its question, checking whether the final report satisfies each criterion and whether its claims are supported by the evidence obtained during the search process. A trajectory is adopted only if its criteria-based score passes a predefined threshold; trajectories that miss key aspects of the question, rely on unsupported claims, or fail to complete the report are discarded. Because the criteria are derived from evidence discovered during question construction, this filtering provides question-specific supervision for open-ended tasks that lack a single reference answer. The retained trajectories are used for supervised fine-tuning on search tasks, teaching the model to plan searches, gather and verify evidence, and synthesize it into comprehensive, evidence-grounded reports.

\subsubsection{Economically Valuable and Real-World Tasks}
\label{sec:ale-tasks}

Training agents for real-world tasks is constrained less by the model than by the supply of high-quality tasks.
A useful task must execute in a real environment, admit an objective check on its outcome, and mirror professional work as it is actually practiced.
We therefore synthesize training environments at scale, then collect and select trajectories in them with strong teacher models.

\noindent\textbf{Benchmark and Motivation.}
Agents' Last Exam (ALE)~\citep{sun2026agentsexam} evaluates agents on long-horizon, economically valuable real-world tasks with verifiable outcomes.
Its tasks span 55 subdomains across 13 industry clusters, covering nonphysical industries defined with reference to O*NET/SOC 2018, the U.S. federal occupational taxonomy.
Each task supplies real input data and a hidden reference.
After the agent finishes, a deterministic grader compares its deliverables against the reference and returns a normalized $[0,1]$ score, without any LLM judge.
Compared with the data pipelines above, ALE complements the training distribution with generalist agent behavior on broad professional workflows, where tasks are built around the specialized software used in actual professional practice, require file and data deliverables produced over multiple steps, and are judged by acceptance criteria from real production settings.
We evaluate general agentic capability on ALE-CLI, the CLI task subset of ALE.

\noindent\textbf{Task Environment Synthesis and Scaling.}
Training tasks originate from a manual cold start of a few dozen high-quality tasks.
Through \emph{controlled evolution}, these seeds have grown into an environment pool of more than one thousand professional workflow environments.
We pair each seed with retrieved public artifacts such as code, datasets, and scientific workflows.
The seed is then varied along at least two substantive axes drawn from the task object, its data, the required operations, workflow dependencies, and the deliverables and grading.
Every seed anchors a broad population of descendants.
A descendant may surpass its ancestor in scientific validity, fidelity to real professional work, and internal consistency.
Difficulty and coverage rise with the population.
Descendants extend well beyond the scope of the seed set, and diversity grows with every round.
Construction and verification are assigned to distinct agents in every evolution round.
A builder agent assembles each candidate package together with its hidden reference and grader.
An independent reviewer agent audits the finished package without seeing the builder's reasoning.
A judger agent compares the candidate against its ancestor and rejects descendants that carry material defects, so only sound descendants are admitted.
Qualified descendants serve as the seeds of the next round, and the environment pool compounds.
The separation is deliberate. Bootstrapping, in which a single model both poses a task and certifies its answer, would corrupt the reward signal.
The hidden reference is produced by executing the reference workflow on the final inputs, which doubles as a solvability check in the target environment.
The grader accepts outputs that match the reference and rejects perturbed ones.
Each task is delivered with an isolated sandbox that reproduces its production context and the professional software the task assumes, with pinned tool versions and a time budget set per task.
Solver and evaluator environments are kept separate, so the agent's execution environment cannot influence grading.

\noindent\textbf{Trajectory Collection and Selection.}
We run each task in its own sandbox and collect trajectories from strong teacher models, with Claude Code as the harness.
Many tasks run long enough to exceed the context window, and the harness then compresses the earlier interaction history and continues the task from the compressed context.
We keep every segment of these trajectories, and the model learns to work under both a full context and a compressed one.
Scoring runs in a separate sandbox that receives the hidden reference only after the attempt has finished, keeping grading independent of the collection process and free of answer leakage.
For each task we select the attempt with the highest score, and we also retain a small share of failed attempts that worked actively on the problem and produced substantive deliverables.
We balance the mixture across tasks and domains, with caps on the share of any single family, of any single domain, and of failed attempts.
The final training set amounts to several hundred million supervised tokens.

\subsection{Unified Training across Domains, Modes, and Harnesses}
\label{sec:unified_training}


Multi-domain SFT combines task-oriented demonstrations with protocol-focused cold-start examples collected across compatible harnesses.
Unified outcome--process post-training then assigns outcome-only, teacher-derived process-only, or mixed supervision at the sample level.

\noindent\textbf{Stage I: Multi-Domain Supervised Fine-Tuning.}
We collect demonstrations under multiple compatible harnesses, including Claude Code, Codex, and internal harnesses.
Harness diversity provides task-preserving variation in tool interfaces, context management, and observations, while allowing each teacher to contribute trajectories from configurations suited to its strengths.
We filter demonstrations using domain-specific checks appropriate to their supervision scope, then canonicalize tool calls and observations while retaining harness metadata.

The resulting data serve two complementary purposes.
Task-oriented demonstrations establish initial problem-solving competence in domains with economical outcome verification.
For machine learning engineering and agent orchestration, protocol-focused cold-start examples primarily establish interaction competence, including tool use, subagent dispatch, and message routing.
The main transfer of domain-specific expertise in these domains occurs through on-policy teacher feedback in Stage II.

For a trajectory $\tau$, let $y_t$ denote the token at position $t$ and $c_t$ its preceding context, including observations and harness metadata.
Let $m_t\in\{0,1\}$ indicate whether $y_t$ is a model-generated target token, and let $N_\tau>0$ denote the number of such tokens.
Across both stages, tool outputs, environment feedback, and harness-injected text are retained as context but excluded from the loss.
Both SFT subsets use the same token-normalized negative log-likelihood:
\begin{equation}
\ell_\theta(\tau) = -\frac{1}{N_\tau}\sum_t m_t\log\pi_\theta(y_t\mid c_t).
\end{equation}

We use temperature-scaled sampling over length-and-turn buckets.
Let $q_{\mathrm{task}}$ and $q_{\mathrm{proto}}$ denote the resulting sampling distributions within the task-oriented and protocol-focused subsets, respectively.
Both distributions retain domain and harness annotations.
Let $\rho\in[0,1]$ denote the total sampling mass assigned to protocol-focused examples.
The SFT objective is
\begin{equation}
\label{eq:unified_sft}
\mathcal{L}_{\mathrm{SFT}}(\theta) = (1-\rho)\mathbb{E}_{q_{\mathrm{task}}}[\ell_\theta] + \rho\mathbb{E}_{q_{\mathrm{proto}}}[\ell_\theta].
\end{equation}
The subsets differ in demonstration content, while sharing the same loss and token mask.

\noindent\textbf{Stage II: Unified Outcome--Process Post-Training.}
Starting from the SFT-initialized policy, we jointly optimize outcome-based RL~\citep{deepseekv32} and multi-teacher on-policy distillation (OPD)~\citep{lu2025onpolicydistillation} on student-generated trajectories.
Let $x=(d,h,\tau)$ denote a student rollout $\tau$ tagged with its domain $d$ and harness $h$.
The sample-level objective is
\begin{equation}
\label{eq:unified_post_training}
\mathcal{L}_{\mathrm{post}}(x) = \lambda_{\mathrm{out}}(x)\mathcal{L}_{\mathrm{RL}}(x) + \lambda_{\mathrm{proc}}(x)\mathcal{L}_{\mathrm{OPD}}(x).
\end{equation}
All loss terms are evaluated under the student policy $\pi_\theta$.
The nonnegative weights reflect the sample's interaction profile and the availability and cost of each signal, selecting outcome-only, process-only, or mixed supervision within a shared training loop.

\emph{Outcome-only supervision} sets $\lambda_{\mathrm{out}}(x)>0$ and $\lambda_{\mathrm{proc}}(x)=0$, yielding mixed-domain RL where outcome verification is economical.
Let $V_d(x)$ denote the domain-specific outcome score, and let $P_{\mathrm{fmt}}(x)$ and $P_{\mathrm{proto}}(x)$ denote nonnegative, calibrated penalties for format and protocol violations.
The reward is
\begin{equation}
\label{eq:unified_outcome}
R_{\mathrm{out}}(x) = V_d(x)-P_{\mathrm{fmt}}(x)-P_{\mathrm{proto}}(x).
\end{equation}
Depending on the domain, this score reflects constraint satisfaction, reference agreement, hidden-test performance, or a native environment metric.
These checks assess task outcomes without establishing the correctness of every intermediate reasoning step.

\emph{Process-only supervision} sets $\lambda_{\mathrm{out}}(x)=0$ and $\lambda_{\mathrm{proc}}(x)>0$, yielding multi-teacher OPD.
The domain tag selects teacher $\mathrm{T}_d$.
Let $p_t$ and $p_t^{\mathrm{T}}$ denote the next-token distributions of the student and the selected teacher, respectively, conditioned on the same student-visited context $c_t$.
Using the shared token mask, the process objective targets reverse KL:
\begin{equation}
\label{eq:unified_opd}
\mathcal{L}_{\mathrm{OPD}}(x) = \frac{1}{N_\tau}\sum_t m_t D_{\mathrm{KL}}(p_t\,\|\,p_t^{\mathrm{T}}).
\end{equation}
Here, process supervision refers to token-level teacher feedback, not a separately trained process reward model.
For machine learning engineering and agent orchestration, dense teacher feedback can supervise available trajectory segments before expensive terminal outcomes arrive.
The choice of this signal is budget-dependent and does not exclude these domains from joint optimization~\citep{kimik3}.

\emph{Mixed supervision} assigns positive weights to both terms on the same student trajectory, combining outcome-driven improvement with teacher guidance for capability retention.
Instruction following and long-context reasoning are examples eligible for both signals~\citep{nemotroncascade2}.

Across supervision modes, student rollouts use compatible harnesses and the shared trajectory format~\citep{glm5}.
Outcome checks use the executed environment state, while teachers score the actual student prefix with matching tool semantics and harness context.
Teachers run as independent prefill services and are queried only when process supervision is active; scoring overlaps with sampling where possible. For agent orchestration task, the student acts as both orchestrator and subagent; a report is supervised in its sender's trajectory and treated as an observation by its receiver.

To accommodate heterogeneous rollout durations, the shared loop uses asynchronous generation and bounded-staleness partial rollouts~\citep{areal, kimik3}.
Outcome advantages are normalized within prompt groups and balanced across domains.
Adaptive sampling tracks domain progress, while response budgets constrain length inflation.
Outcome-based dynamic sampling handles negligible reward variation~\citep{dapo} without discarding process-only samples or useful teacher feedback in mixed samples.
Cross-domain evaluation monitors interference and teacher incompatibility, while held-out-harness evaluation assesses sensitivity to scaffold conventions.

\subsection{Agent Training Infrastructure}
Agentic Orchestration expands exploration across concurrent investigations, while Agentic Verification revisits evidence and redirects subsequent actions. In engineering and scientific tasks, these behaviors require files, processes, and intermediate results to remain available as agents test alternatives and revise their approaches. Their logical progress is continuous, but their demand for compute is intermittent. The infrastructure challenge is therefore to preserve many evolving execution contexts without keeping every environment fully resident throughout its lifetime. We build on PolicyVisor (pVisor), an agent-native sandbox framework developed by our infrastructure team, to address this challenge. It provides the default sandbox environments for the agentic rollouts in the experiments section, with MLE-Bench-Lite running on dedicated GPU hosts. Its central principle is to make file and network interactions visible to the host for access control and resource management.

\noindent\textbf{pVisor.} To support agent training and future deployment at larger scales, pVisor combines on-demand environment loading with memory management across execution phases. Shared software dependencies are loaded as needed, while fixed environment versions provide consistent starting points for repeated attempts. Each workspace retains the code and experimental outputs produced along its execution path, allowing subsequent tool calls to inspect results, run additional checks, and refine a solution. Access policies control file and network operations, and workspace changes are retained for review. These mechanisms preserve the execution context and artifacts needed for continued exploration and task-specific evaluation.

While an agent waits for a model response, its environment may continue to occupy substantial memory. pVisor can pause execution, offload memory, and resume from the same point. An experimental integration allows the host to advance model requests during these waits and resume the environment to deliver the response, preserving the agent's existing tools and control flow. This separates the lifetime of a rollout from continuous residency of its full execution environment. Preliminary static tests, with one run per condition, showed approximately 80\% lower resident memory across the VM and supporting processes after offload than in a running baseline with the same application working set. Agent workloads vary widely in memory demand, waiting duration, and concurrent tool activity; actual savings also depend on the cost of managing requests and restoring execution. The model determines which investigations to pursue and when to revise them; the runtime manages the resources needed to maintain those investigations across active and waiting phases.
\section{Experiment}

\subsection{Evaluation Setting}  

\textbf{Search Benchmarks.} To evaluate long-horizon information seeking and deep research capabilities, we evaluate \ProjectName{} on four search benchmarks. \textbf{GAIA}~\citep{gaia} consists of real-world assistant questions that require multi-step reasoning, web browsing, and tool use. \textbf{BrowseComp}~\citep{browsecomp} contains challenging questions whose answers are hard to find but easy to verify, requiring persistent and creative browsing over a large number of web pages. \textbf{DeepResearch Bench II}~\citep{li2026deepresearchbenchiidiagnosing} evaluates deep research agents on report generation tasks, scoring the generated reports against rubrics derived from expert-written reports. \textbf{XBench-DS-2510}~\citep{xbench} is the October 2025 release of the xbench DeepSearch suite, assessing deep search abilities on profession-aligned, real-world questions. For all search benchmarks, the agent is equipped with three tools: a \textit{search} tool that retrieves web pages and returns the top-50 results per query; a \textit{visit} tool that fetches webpage content and summarizes it with a dedicated summarization model to extract task-relevant information; and a \textit{code} tool that executes Python scripts in a remote sandbox to support complex computation and logical reasoning. For long-horizon tasks such as BrowseComp, we further apply a context management strategy: once the agent exceeds 300 tool calls or reaches the context limit of the model, the accumulated context is discarded and the agent is rerun on the task from scratch, with at most 5 rounds allowed.
For Agentic Orchestration evaluation, we augment the search setup with the \texttt{spawn\_subagent} tool for the main agent. When a subagent exceeds 50 turns, it is required via a prompt to summarize its findings and return the result to the main agent.

\textbf{Engineering Benchmarks.} For \textbf{SWE-Bench-Pro}~\citep{deng2025swebenchpro} and \textbf{SWE-Bench-Multilingual}~\citep{yang2025swesmith}, we evaluate \ProjectName{} using Mini-SWE-Agent~\citep{swe-agent}. SWE-Bench-Pro targets complex, enterprise-oriented repository changes, while SWE-Bench-Multilingual assesses issue resolution across nine programming languages. Both benchmarks evaluate whether generated patches resolve the target issue while preserving existing functionality. To mitigate benchmark leakage and evaluation hacking, we restrict access to Git history that could reveal reference solutions and disable network access throughout evaluation. Besides, we evaluate \ProjectName{} on \textbf{SciCode}~\citep{scicode} and \textbf{MLE-Bench-Lite}~\citep{mlebench} using the same official protocols as A1. For SciCode, we provide scientist-annotated background for each sequential subproblem and report pass@1 across the 288 test subproblems, with correctness requiring all associated hidden unit tests to pass. MLE-Bench-Lite evaluates autonomous ML engineering across 22 Kaggle competitions, requiring agents to explore data, train models, and generate submissions from only the dataset and task description. Submissions are graded against the original competition leaderboards using the official MLE-Bench grader. We report the fraction earning at least a bronze medal, averaged over three seeds. Each task runs in isolation on a dedicated H200 GPU with a 12-hour wall-clock budget.

\textbf{Scientific Research Benchmarks.} To evaluate scientific research capabilities, we select six representative benchmarks: \textbf{FrontierScience-Olympiad (FS-O)}~\cite{wang2026frontierscience} covers Olympiad-level problems in physics, chemistry, and biology, assessing advanced scientific reasoning across multiple disciplines. \textbf{FrontierScience-Research (FS-R)}~\cite{wang2026frontierscience} comprises PhD-level research subtasks in physics, chemistry, and biology, evaluating the ability to solve challenging scientific research problems. \textbf{HLE w/ tools}~\cite{2026hle} features expert-level academic questions across diverse disciplines, challenging models' advanced knowledge, complex reasoning, and tool-use capabilities. \textbf{HiPhO}~\cite{yu2025hiphofarmllmshumans} is the first dedicated physics Olympiad benchmark, encompassing 13 physics competitions from 2024--2025 with fine-grained, human-aligned grading to evaluate multimodal physical reasoning. \textbf{CMT-Benchmark (CMT)}~\cite{pan2026cmtbenchmarkbenchmarkcondensedmatter} focuses on expert-level problems in condensed matter theory, requiring specialized scientific knowledge, rigorous analytical reasoning, and advanced computation. \textbf{CritPt}~\cite{zhu2026probingcriticalpointcritpt} presents frontier-level physics research challenges with computationally verifiable solutions, assessing complex scientific reasoning and research-oriented problem solving. For evaluation metrics, we report the average competition score across all 13 physics Olympiads for HiPhO and accuracy for the remaining five benchmarks. For CritPt, we submit model-generated solutions to Artificial Analysis for official evaluation and report the resulting accuracy.

\textbf{General Benchmarks.} To assess general capabilities, we evaluate \ProjectName{} on three complementary benchmarks: \textbf{AA-LCR}~\citep{artificialanalysis2025lcr} for long-context reasoning, \textbf{IFBench}~\citep{pyatkin2025ifbench} for instruction following, and \textbf{ALE-CLI}~\citep{sun2026agentsexam} for general agentic capability on economically valuable real-world tasks. AA-LCR comprises 100 multi-document reasoning questions that require models to locate and cross-reference specific facts across large document collections. Each question is scored by an LLM judge against a reference answer, and we report accuracy averaged over three independent runs. This benchmark probes whether a model can maintain coherent retrieval and reasoning over extended contexts rather than relying on parametric knowledge alone. IFBench evaluates the ability to follow complex, compositional instructions that combine multiple simultaneous constraints on open-ended generation tasks. Each of the 294 test prompts specifies a set of verifiable constraints, and correctness requires satisfying all of them simultaneously. We report strict prompt-level accuracy for IFBench. ALE-CLI comprises 105 tasks across 12 domains, and we evaluate it using Claude Code, with every run executed independently in a clean sandbox under each task's default configuration. Tasks are assigned to sandbox images according to their workload requirements, covering general-purpose, desktop, R, scientific, engineering, and GUI environments, with each task running in a CPU-only sandbox with an eight-hour timeout, typically using 4 CPUs and 16 GiB of memory. The maximum output and thinking-token budgets are set to 64K and 32K tokens, respectively. Agent-generated artifacts are evaluated using the official task-specific graders, and we report the pass rate, where a task is counted as passed only when it receives full score.
Pinchbench~\citep{pinchbench2026}. We evaluate \ProjectName{} on the 23 tasks of PinchBench-v1 using OpenClaw. Following the official repeated-evaluation protocol, each task is evaluated over five runs in the isolated sandbox. The context window and maximum output length are set to 262K and 65K tokens, respectively, with task-dependent timeouts ranging from 3 to 15 minutes. DeepSeek-V4-Flash is used for LLM-based grading.

\begin{table}[t]
    \centering
    \footnotesize
    \setlength{\tabcolsep}{1.9pt}
    \caption{Comparison between \ProjectName~and 35B/T-level models. For 35B model, we compare with recently released open-source 35B models. For T-level models, we compare with Kimi-K3~\citep{kimik3}, GLM-5.3~\citep{GLM-5.3}, and GPT-5.6-sol~\citep{GPT-5.6}. To ensure a fair comparison, we report the results from their original technical reports. If a model does not report the corresponding benchmark results, we evaluate it using the same evaluation protocol as our model. \underline{underline} means the best result for 35B parameters, and \textbf{Bold} means the best overall result. Note that GLM-5.3 is text-only and thus not applicable to OmniDocBench.}
    \label{tab:main_results}

    \begin{tabular}{@{}lcccccccc@{}}
    \toprule
    & \multicolumn{3}{c}{\textbf{Comparable Models (~35B)}} & \multicolumn{3}{c}{\textbf{Larger-scale Models}} \\
    \cmidrule(lr){2-4} \cmidrule(lr){5-7}
    & \multicolumn{1}{l}{\textbf{\ProjectName}} & \multicolumn{1}{l}{\textbf{Agents-A1}} & \multicolumn{1}{l}{\textbf{Ornith-1.5}} & \multicolumn{1}{l}{\textbf{Kimi-K3}} & \multicolumn{1}{l}{\textbf{GLM-5.3}} & \multicolumn{1}{l}{\textbf{GPT-5.6-sol}} \\
    \midrule
    \textbf{Long-horizon Search} & \multicolumn{1}{l}{} & \multicolumn{1}{l}{} & \multicolumn{1}{l}{} & \multicolumn{1}{l}{} & \multicolumn{1}{l}{} & \multicolumn{1}{l}{} & \multicolumn{1}{l}{} & \multicolumn{1}{l}{} \\
    \midrule
    GAIA~\citep{gaia} & 95.1 & \textbf{96.0} & 77.6 & 92.2 & 92.2 & 88.3 \\
    BrowseComp~\citep{browsecomp} & \underline{81.7} & 75.5 & 67.6 & \textbf{91.2} & 76.2 & 90.4 \\
    DeepResearch-Bench2~\citep{li2026deepresearchbenchiidiagnosing} & 43.7 & 37.5 & \underline{48.6} & 52.4 & 55.7 & \textbf{57.9} \\
    XBench-DS-2510~\citep{xbench} & \textbf{92.0} & 86.0 & 80.0 & 85.0 & 84.0 & 90.0 \\
    \midrule
    \textbf{Engineering Tasks} & & & & & & \\
    \midrule
    MLE-Bench-Lite~\citep{mlebench} & \underline{65.2} & 43.9 & 39.4 & 72.7 & 72.7 & \textbf{77.3} \\
    SciCode~\citep{scicode} & \underline{44.3} & 44.3 & 39.5 & 58.7 & \textbf{59.0} & 56.1 \\
    SWE-Bench Pro~\citep{deng2025swebenchpro} & \underline{61.2} & 43.6 & 59.6 & 63.3 & \textbf{64.6} & \textbf{64.6} \\
    SWE-Bench Multilingual~\citep{yang2025swesmith} & \underline{72.7} & 58.7 & 71.4 & 80.8 & \textbf{81.3} & 74.1 \\
    \midrule
    \textbf{Instruction Following} & & & & & & \\
    \midrule
    IFbench~\citep{pyatkin2025ifbench} & \textbf{81.6} & 80.6 & 44.6 & 72.1 & 53.4 & 72.7 \\
    AA-LCR~\citep{artificialanalysis2025lcr} & \underline{81.0} & 74.2 & 73.2 & \textbf{88.7} & 79.7 & 84.0 \\
    \midrule
    \textbf{Scientific Research} & & & & & & \\
    \midrule
    FS-O~\citep{wang2026frontierscience} & \textbf{81.0} & 79.0 & 70.0 & 69.0 & 67.0 & 73.0 \\
    FS-R~\citep{wang2026frontierscience} & \textbf{48.3} & 40.0 & 6.7 & 21.7 & 16.7 & 26.7 \\
    HiPhO~\citep{yu2025hiphofarmllmshumans} & 45.3 & \textbf{46.4} & 39.5 & 42.4 & 39.7 & 46.2 \\
    HLE w/ tools~\citep{2026hle} & \underline{55.1} & 47.6 & 33.4 & 56.0 & 62.5 & \textbf{64.5} \\
    CMT~\citep{pan2026cmtbenchmarkbenchmarkcondensedmatter} & \textbf{80.0} & 72.0 & 26.0 & 34.0 & 40.0 & 50.0 \\
    CritPt~\citep{zhu2026probingcriticalpointcritpt} & \underline{20.0} & 4.3 & 2.9 & 23.4 & 19.1 & \textbf{32.3} \\
    \midrule
    \textbf{General Agentic Tasks} & & & & & & \\
    \midrule
    Agents' Last Exam (ALE-CLI)~\citep{sun2026agentsexam} & \underline{19.1} & 10.2 & 14.3 & 27.6 & 28.5 & \textbf{28.6} \\
    VitaBench~\citep{he2025vitabench} & \underline{45.4} & 38.8 & 44.6 & \textbf{59.0} & 58.7 & 46.8 \\
    PinchBench~\citep{pinchbench2026} & \textbf{90.0} & 76.9 & 88.6 & 88.2 & 86.6 & 87.3 \\
    \midrule
    \textbf{Scientific Agentic Tasks} & & & & & & \\
    \midrule
    ChemCOTBench~\citep{li2025beyond} & \underline{66.5} & 64.0 & 53.8 & \textbf{82.5} & 78.3 & 77.9 \\
    MatTools~\citep{MatTools} & 42.6 & \underline{47.1} & 38.6 & 24.3 & \textbf{53.2} & 45.6 \\
    OmniDocBench~\citep{ouyang2024omnidocbenchbenchmarkingdiversepdf} & \textbf{91.5} & 89.5 & 89.6 & 91.1 & - & 85.8 \\
    \bottomrule
    \end{tabular}
\end{table}

\textbf{Scientific Agentic Tasks.} To evaluate scientific agentic capabilities, we evaluate \ProjectName{} on three complementary benchmarks in Table~\ref{tab:main_results}: \textbf{ChemCoTBench}~\citep{li2025beyond} for chemical reasoning, \textbf{MatTools}~\citep{MatTools} for materials science tool use, and \textbf{OmniDocBench}~\citep{ouyang2024omnidocbenchbenchmarkingdiversepdf} for document parsing. For ChemCoTBench, we use a frozen evaluation set of 1,495 examples covering 19 subtasks in molecular understanding, molecular editing, molecular property optimization, and chemical reactions. Models receive the original queries without additional demonstrations or tools, with a maximum output length of 65,536 tokens including reasoning and the final answer. We score final answers using a frozen adapter to the released chemical evaluation functions and report the unweighted mean of the 19 subtask metrics. For MatTools, we follow the official real-world tool-usage evaluation: models generate Python code using \texttt{pymatgen} to solve materials property calculation problems, and the code is executed in a Docker sandbox and checked using the provided verification scripts. We report the task success rate across 138 subtasks from 49 problems. For OmniDocBench, we follow the official end-to-end evaluation, matching page-level Markdown predictions against annotated document components. We report the official overall score, averaging text similarity based on normalized edit distance, table TEDS, and formula CDM on a common percentage scale.

\subsection{Results and Observations} 
\textbf{Performance on Scientific Domain.} As shown in Table~\ref{tab:science_teacher_comparison}, the Science-enhanced Teacher consistently outperforms the Qwen3.5-35B-A3B baseline across all six scientific benchmarks, with particularly substantial gains on challenging research-level tasks. On Olympiad-level benchmarks, performance improves from 64.5 to 89.0 on FS-O and from 37.0 to 49.7 on HiPhO, demonstrating stronger scientific reasoning capabilities. More notably, the teacher achieves remarkable improvements on research-level benchmarks, with scores increasing from 2.5 to 70.0 on FS-R, 20.0 to 84.0 on CMT, and 0.9 to 25.7 on CritPt, highlighting its enhanced ability to tackle complex scientific research problems requiring advanced reasoning and tool-assisted computation. On HLE with tools, accuracy further increases from 47.4 to 56.8, demonstrating improved performance on diverse expert-level academic challenges. Overall, these results highlight the effectiveness of the proposed training pipeline in strengthening scientific reasoning and tool-augmented problem-solving capabilities, with particularly pronounced improvements on high-difficulty scientific research tasks.
\begin{table}[tbp]
\centering
\caption{Performance comparison between Qwen3.5-35B-A3B and the science-enhanced teacher.}
\label{tab:science_teacher_comparison}
\footnotesize
\setlength{\tabcolsep}{8pt}
\renewcommand{\arraystretch}{1.4}
\begin{tabular}{@{}lcccccc@{}}
          \toprule
          \textbf{Model} & \textbf{HLE w/ tools} & \textbf{HiPhO} & \textbf{FS-O} & \textbf{FS-R} & \textbf{CMT} & \textbf{CritPt} \\
          \midrule
          Qwen3.5-35B-A3B
          & 47.4 & 37.0 & 64.5 & 2.5 & 20.0 & 0.9 \\
          Science-enhanced Teacher (SFT)
          & \textbf{56.8}
          & \textbf{49.7}
          & \textbf{89.0}
          & \textbf{70.0}
          & \textbf{84.0}
          & \textbf{25.7}\\
          \bottomrule
      \end{tabular}
\end{table}

\textbf{Performance on Search Domain.} As shown in Table~\ref{tab:search_teacher_comparison}, the Search-enhanced Teacher consistently outperforms the Qwen3.5-35B-A3B baseline across all four search benchmarks. The most pronounced gains appear on tasks that demand multi-step information seeking: accuracy improves from 59.8 to 96.1 on GAIA and from 61.0 to 85.5 on BrowseComp, indicating that the teacher can sustain long-horizon browsing, persistently explore hard-to-find evidence, and verify candidate answers. On XBench-DS-2510, performance further increases from 77.0 to 89.0, demonstrating stronger deep search capabilities on profession-aligned, real-world questions. On DeepResearch Bench II, the score rises from 42.3 to 48.2, suggesting improved ability to synthesize retrieved information into comprehensive research reports. Overall, these results highlight the effectiveness of the proposed training pipeline in strengthening long-horizon search and tool-augmented information seeking.

\begin{table}[tbp]
\centering
\caption{Performance comparison between Qwen3.5-35B-A3B and the search-enhanced teacher.}
\label{tab:search_teacher_comparison}
\footnotesize
\setlength{\tabcolsep}{8pt}
\renewcommand{\arraystretch}{1.4}
\begin{tabular}{@{}lcccc@{}}
          \toprule
          \textbf{Model} & \textbf{GAIA} & \textbf{BrowseComp} & \textbf{DeepResearch Bench II} & \textbf{XBench-DS-2510} \\
          \midrule
          Qwen3.5-35B-A3B
          & 59.8 & 61.0 & 42.3 & 77.0 \\
          Search-enhanced Teacher (SFT)
          & \textbf{96.1}
          & \textbf{85.5}
          & \textbf{48.2}
          & \textbf{89.0} \\
          \bottomrule
      \end{tabular}
\end{table}

\begin{table}[htbp]
\centering
\caption{Performance of Agentic Orchestration on BrowseComp~\citep{browsecomp} and DeepResearch Bench II~\citep{li2026deepresearchbenchiidiagnosing}. The teacher is evaluated with Agentic Orchestration enabled. Bold denotes the best result in each column.}
\label{tab:orchestration_teacher_results}
\footnotesize
\setlength{\tabcolsep}{8pt}
\renewcommand{\arraystretch}{1.4}
\begin{tabular}{@{}lcc@{}}
\toprule
\textbf{Model} & \textbf{BrowseComp} & \textbf{DeepResearch Bench II} \\
\midrule
Qwen3.5-35B-A3B & 61.0 & 42.3 \\
\ProjectName{} w/o Agentic Orchestration & 81.7 & 43.7 \\
\ProjectName{} w/ Agentic Orchestration & \textbf{82.8} & 47.5 \\
Agentic Orchestration Teacher & 81.6 & \textbf{53.6} \\
\bottomrule
\end{tabular}
\end{table}

\noindent\textbf{Performance of the Agentic Orchestration.} As shown in Table~\ref{tab:orchestration_teacher_results}, the Agentic Orchestration teacher achieves 81.6 on BrowseComp and 53.6 on DeepResearch Bench II, compared with 61.0 and 42.3 for Qwen3.5-35B-A3B. For \ProjectName{}, enabling Agentic Orchestration raises BrowseComp accuracy from 81.7 to 82.8 and the DeepResearch Bench II score from 43.7 to 47.5, corresponding to gains of 1.1 percentage points and 3.8 points, respectively. \ProjectName{} with orchestration achieves the highest BrowseComp accuracy among these configurations, while the teacher obtains the highest report-generation score. The larger gain for \ProjectName{} on DeepResearch Bench II is consistent with the role of orchestration in coordinating complementary investigations and integrating their findings into a coherent research report.

\begin{table}[htbp]
\centering
\caption{Performance and token costs of Agentic Verification on BrowseComp~\citep{browsecomp}. Tokens and turns are averaged per task.}
\label{tab:verification_teacher_results}
\footnotesize
\setlength{\tabcolsep}{8pt}
\renewcommand{\arraystretch}{1.4}
\begin{tabular}{@{}lccc@{}}
\toprule
\textbf{Model / Mode} & \textbf{BrowseComp} & \textbf{Output Tokens} & \textbf{Turns} \\
\midrule
Qwen3.5-35B-A3B & 61.0 & 14,851 & 65.22 \\
Agentic Verification teacher (normal mode) & 81.0 & 23,245 & 84.81 \\
Agentic Verification teacher (verify mode) & \textbf{85.5} & 21,962 & 82.14 \\
\bottomrule
\end{tabular}
\end{table}

\noindent\textbf{Performance of the Agentic Verification.} As shown in Table~\ref{tab:verification_teacher_results}, the Verify teacher achieves 81.0 accuracy on BrowseComp in normal mode, compared with 61.0 for Qwen3.5-35B-A3B. Enabling Agentic Verification further improves accuracy to 85.5, a gain of 4.5 percentage points and the highest accuracy among the compared configurations. In paired attempts where both modes produced an answer, the average output token count per task decreases from 23,245 to 21,962, a reduction of 5.52, while the average number of interaction turns decreases from 84.81 to 82.14, a reduction of 3.15. Within this paired comparison, the additional reasoning introduced by verification is therefore accompanied by a net reduction in generation costs. This trend is consistent with the design goal of Agentic Verification: reducing unproductive exploration by actively reviewing evidence, identifying ineffective paths, and adjusting subsequent strategies.

\begin{table}[tbp]
\centering
\caption{Performance comparison between Qwen3.6-35B-A3B and the MLE teacher on MLE-Bench Lite~\cite{mlebench}. Results are reported as mean $\pm$ SEM over three evaluations (\%).}
\label{tab:mle_teacher_results}
\begingroup
\footnotesize
\setlength{\tabcolsep}{8pt}
\renewcommand{\arraystretch}{1.4}
\begin{tabular}{@{}lccccc@{}}
\toprule
\textbf{Model} & \textbf{Med+} & \textbf{Gold} & \textbf{Silver} & \textbf{Bronze} & \textbf{Any Medal} \\
\midrule
Qwen3.6-35B-A3B & $62.12 \pm 4.01$ & $22.73 \pm 2.62$ & $7.58 \pm 4.01$ & $7.58 \pm 1.52$ & $37.88 \pm 5.46$ \\
MLE Teacher (SFT+RL) & $\mathbf{84.85} \pm 3.03$ & $\mathbf{50.00} \pm 2.62$ & $\mathbf{16.67} \pm 1.52$ & $6.06 \pm 1.52$ & $\mathbf{72.73} \pm 2.62$ \\
\bottomrule
\end{tabular}
\endgroup
\end{table}

\noindent\textbf{Performance of the MLE Teacher.} As shown in Table~\ref{tab:mle_teacher_results}, the MLE teacher achieves a medal rate of 72.7\% on MLE-Bench Lite, outperforming its Qwen3.6-35B-A3B base model at 37.9\% by 34.8 percentage points. Its gold medal rate reaches 50.0\%, more than twice that of the base model, indicating gains in both medal coverage and the ability to achieve higher competitive performance. These results show that agentic post-training effectively improves the model's end-to-end machine learning engineering capability.

\textbf{Performance on General Domain.}
\ProjectName{} achieves a strict pass rate of 19.0\% on ALE-CLI, ranking fourth overall but first among the evaluated 35B-class models. It outperforms A1 (10.2\%) and Ornith-1.5 (14.3\%) by 8.8 and 4.7 percentage points, respectively. Although it still trails Kimi-K3 (27.6\%), GLM-5.3 (28.5\%), and GPT-5.6-sol (28.6\%), the results demonstrate strong performance relative to its model scale. Domain-level analysis further shows that its main strengths lie in Business and Finance (28.6\%) and Computing and Mathematics (26.3\%), while Health and Medicine (13.6\%) and Engineering (12.5\%) remain the primary areas for improvement.
With an overall score of 90.0, our 35B model \ProjectName{} ranks first among all evaluated systems, outperforming proprietary models including Kimi-K3 (88.2), GLM-5.3 (86.6), and GPT-5.6-sol (87.3). It excels at coding, file operations, and spreadsheet analysis, scoring 97.6–100.
\section{Open Scientific Discovery via \ProjectName{}: Case Studies}

Real-world scientific research unfolds over long horizons, with reasoning, experimentation, and analysis building on one another. This poses challenges in coordinating interdependent tasks and preventing errors from affecting subsequent steps. \ProjectName~uses agent orchestration to distribute tasks among specialized subagents and verification to check assumptions and intermediate results. Together, these mechanisms highlight the potential of \ProjectName~ to move beyond isolated task execution toward sustained scientific discovery. The following showcases illustrate how these capabilities support scientific research in practice.

\begin{figure}[t]
    \centering
    \includegraphics[width=0.90\linewidth]{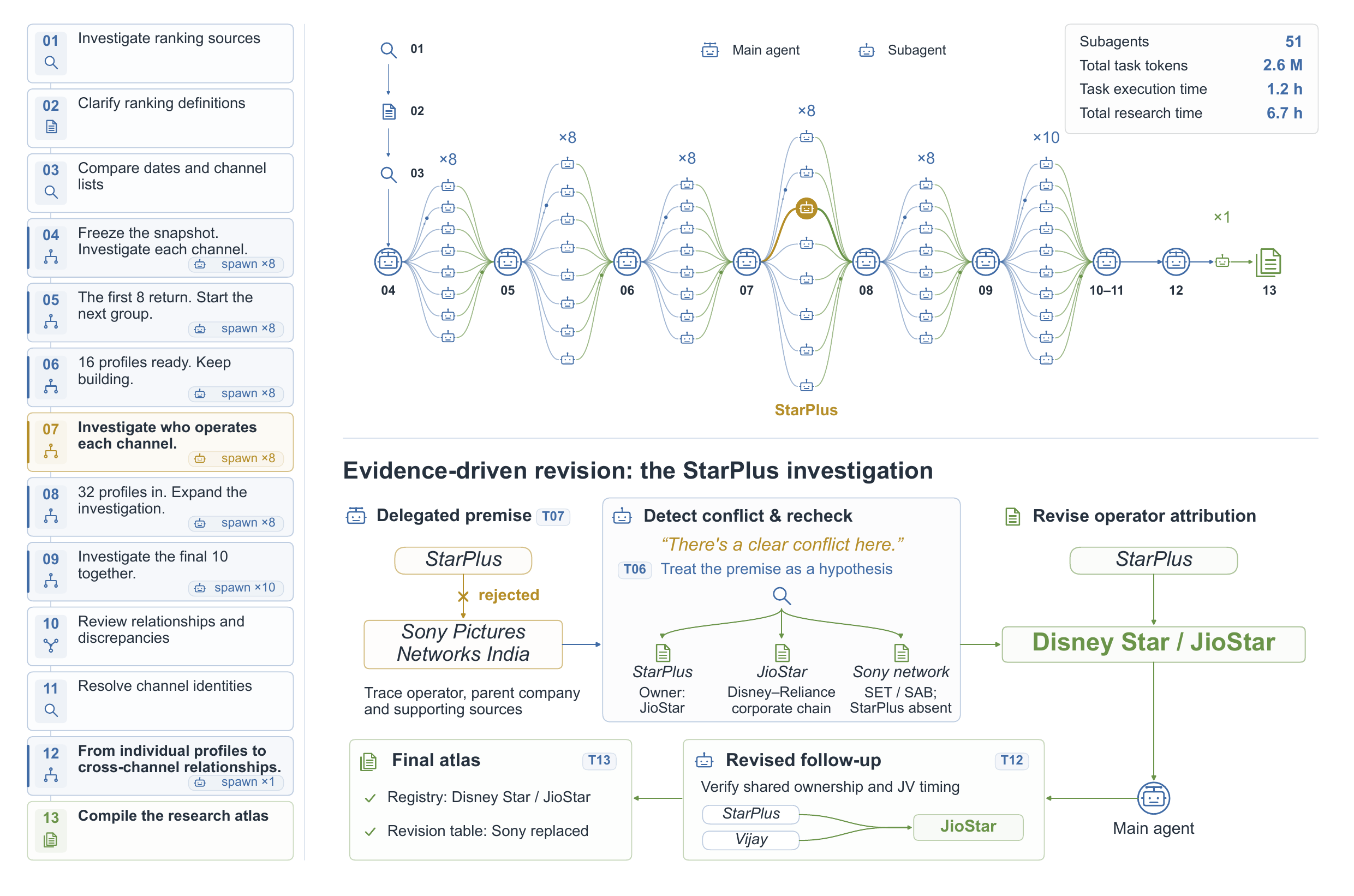}
    \caption{Agentic Orchestration in a 50-channel research atlas. \textbf{Left:} the 13-turn main-agent trajectory, with badges marking delegation and batch sizes. \textbf{Upper right:} parallel subagents investigate 50 channels in six successive batches, after which one further subagent examines cross-channel relationships. \textbf{Lower right:} the \textit{StarPlus} subagent tests the delegated \textit{Sony} attribution against evidence and corrects it to \textit{Disney Star / JioStar}; the main agent uses this correction in its subsequent relationship assignment and final atlas. Task execution time measures the end-to-end elapsed duration; total research time sums the execution durations of all subagents.}
    \label{fig:orchestration-showcase}
\end{figure}

\begin{figure}[t]
    \centering
    \includegraphics[width=0.90\linewidth]{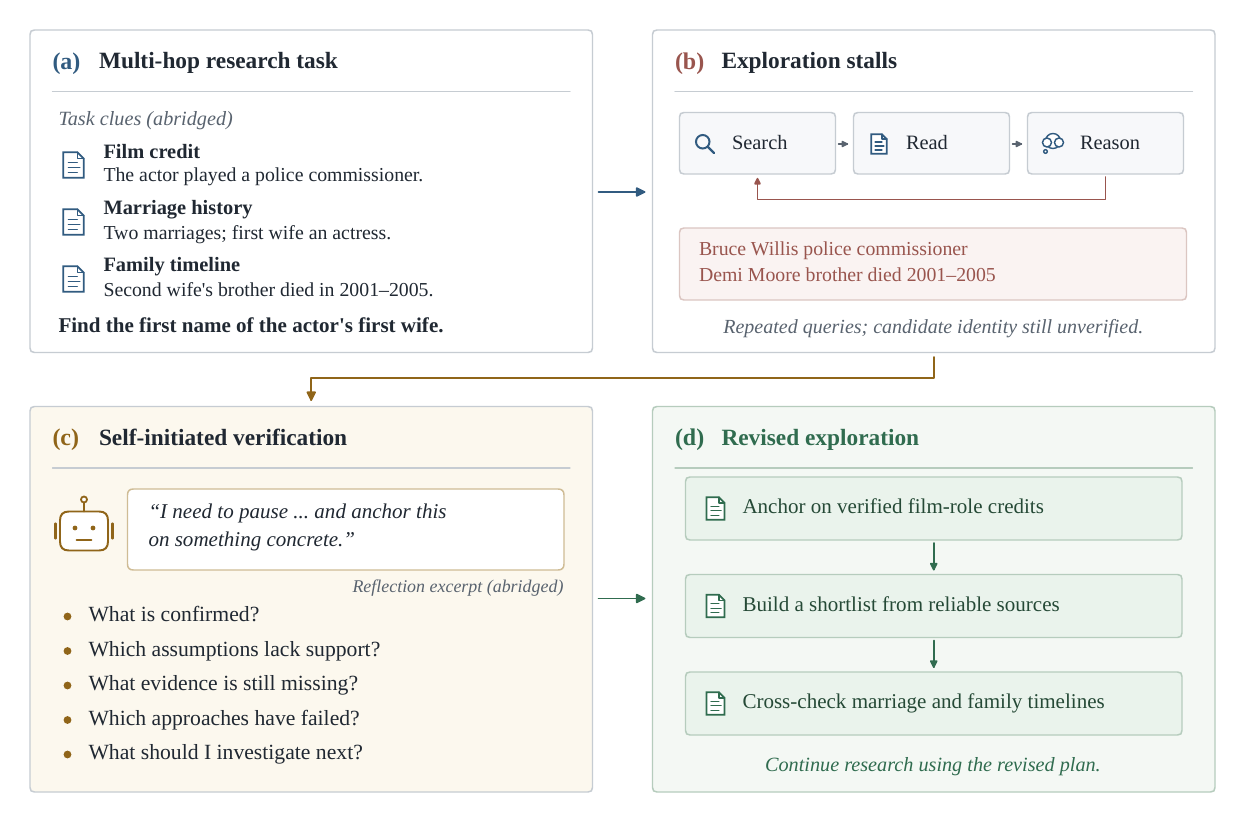}
    \caption{
    Agentic Verification in a multi-hop BrowseComp task.
    (a) The task identifies an actor's first wife through clues about film roles, marriages, and family history.
    (b) Repeated searches fail to establish a reliable candidate.
    (c) The model autonomously initiates verification, identifying unsupported assumptions and evidence gaps.
    (d) It revises its search strategy to verify film credits and marital relationships.
    Red, amber, and green indicate stalled exploration, self-initiated verification, and revised exploration, respectively.
    }
    \label{fig:showcase_verify}
\end{figure}

\subsection{Foundational Capabilities}

\subsubsection{Agentic Orchestration}

To demonstrate how \ProjectName{} coordinates complex research tasks, we consider a large-scale investigation of the world's 50 most-viewed YouTube channels, covering channel identities, ownership structures, and commercial relationships. The task requires extensive evidence gathering and cross-entity analysis, where findings from individual investigations may affect subsequent research decisions. \ProjectName{} decomposes the task into 50 channel-specific investigations, dispatched in six batches of parallel subagents (Fig.~\ref{fig:orchestration-showcase}). Each subagent independently gathers evidence, while the main agent coordinates their findings and identifies questions requiring further investigation.

Rather than simply aggregating results, the main agent dynamically adjusts the research process based on intermediate findings. For example, it initiates follow-up investigations into relationships among channels sharing commercial partners, distinguishing partnership agreements from common ownership. Subagents also challenge initial assumptions when conflicting evidence emerges. In the \textit{StarPlus} investigation, a subagent corrects the initially assigned operator from \textit{Sony Pictures Networks India} to \textit{Disney Star / JioStar}, and this correction is incorporated into subsequent analysis. The main agent ultimately integrates the findings into a structured report covering channel profiles, ownership relationships, and content-supply models. This case illustrates how agentic orchestration extends beyond parallel task execution, enabling research findings to guide subsequent investigations and refine the overall research process.

\begin{figure*}[t]
    \centering
    \includegraphics[width=\textwidth]{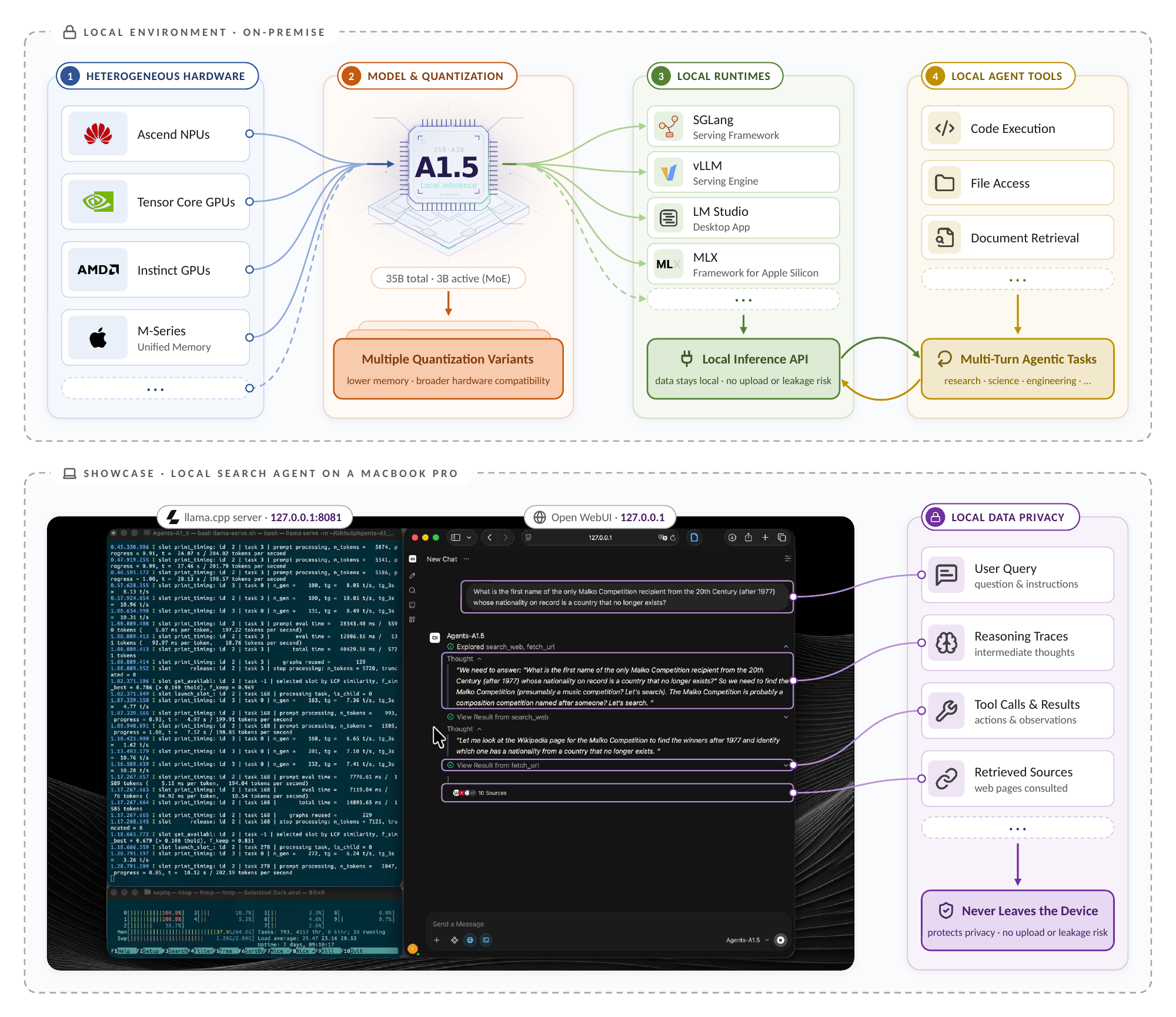}
    \caption{
        \textbf{Local deployment of \ProjectName{}.}
          Quantized variants of \ProjectName{} (MoE with 35B total and 3B active parameters) run
          on heterogeneous hardware, are served by local runtimes through a local
          inference API, and carry out multi-turn scientific and engineering tasks with
          local agent tools. All components stay within the local environment (dashed
          boundary), enabling a fully on-premise agentic workflow.
    }
    \label{fig:local_deployment}
\end{figure*}

\subsubsection{Agentic Verification}

To examine Agentic Verification in long-horizon exploration, we compare the trajectories of the trained model and a baseline on a multi-hop BrowseComp question. The task asks for the first name of an actor's first wife, based on clues about his portrayal of a police commissioner in a film, his two marriages, and his second wife's family history. Solving this question requires cross-checking multiple constraints, where partial matches can lead to incorrect conclusions. As shown in Fig.~\ref{fig:showcase_verify}, Agentic Verification allows the model to assess its progress and revise its strategy when necessary.

The model initially explored broad candidate matches but autonomously initiated verification at turn 20. It recognized that the available evidence supported only parts of the required constraints and that repeated searches had yielded limited progress. Based on this assessment, the model shifted to targeted searches for film credits and marital relationships, ultimately producing \textit{Madhumati} at turn 30, which the case annotation accepts as a spelling variant of the reference answer \textit{Madhumalti}.

In contrast, the baseline continued broad searches and eventually selected Tom Selleck despite unresolved evidence, using a television role to satisfy the film-role constraint. It returned the incorrect answer \textit{Jacqueline} after 48 turns. This comparison illustrates how Agentic Verification can redirect a stalled exploration by identifying evidence gaps and using intermediate assessments to guide subsequent actions.

\begin{figure}[t!]
    \centering
    \includegraphics[width=0.8\linewidth]{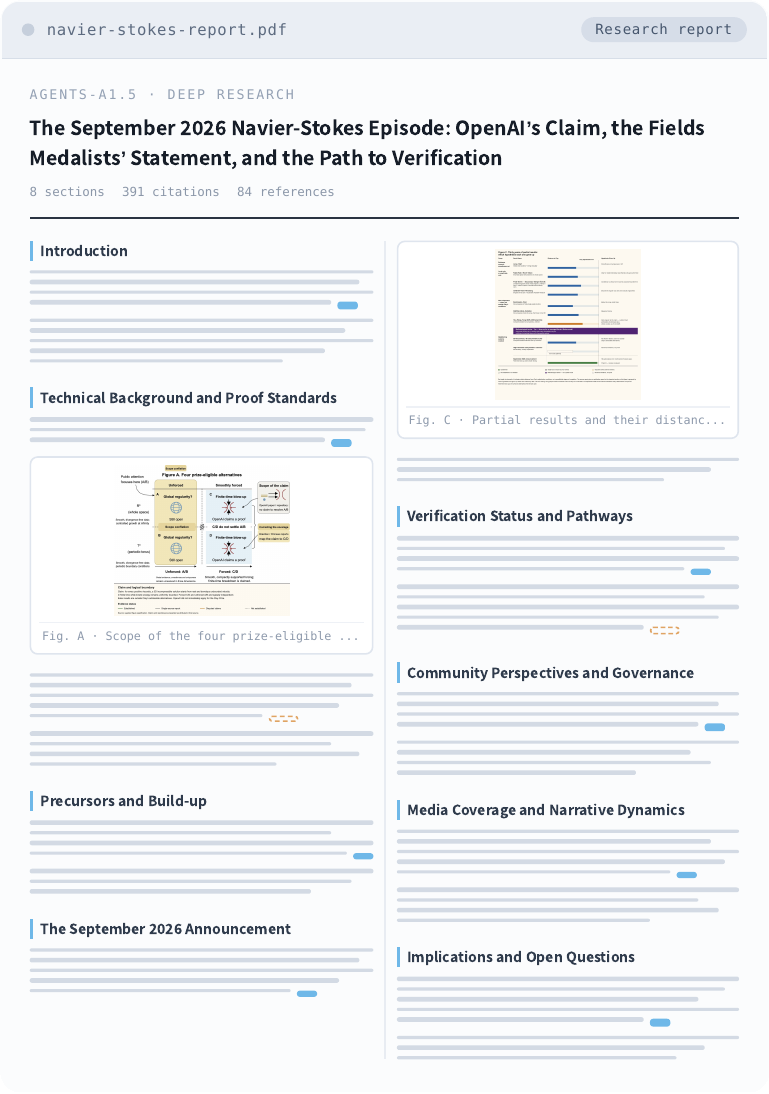}
    \caption{Illustrative preview of a DeepResearch report generated \ProjectName{}}
    \label{fig:showcase_deepresearch}
\end{figure}

\subsubsection{Local Automated Deployment}
\ProjectName{} is designed to support flexible deployment across diverse local computing environments, ranging from data-center accelerators to consumer-grade workstations. The 35B-A3B model can be served on heterogeneous hardware including domestic NPUs such as Huawei Ascend 910B/910C, mainstream NVIDIA GPUs such as A100, H100, H200, B200, and RTX-series cards, AMD GPUs and Instinct accelerators such as MI300X, and Apple Silicon machines including M-series chips with unified memory. To further reduce memory requirements and broaden hardware compatibility, the model provides multiple quantization variants, including FP8 and 4-bit formats. These options make it feasible to run \ProjectName{} on local devices with limited accelerator memory while preserving practical inference quality and throughput.

In addition to local model serving, \ProjectName{} can be integrated with local toolchains to enable fully on-premise agentic workflows, as illustrated in Fig.~\ref{fig:local_deployment}. Through deployment tools and runtimes such as LM Studio, llama.cpp, and Apple MLX, users can launch the model with minimal configuration and expose it through local inference APIs. Code execution, file access, document retrieval, and other agent tools can likewise be executed in the user's local environment. This allows the agentic model to complete multi-step scientific and engineering tasks without relying on external cloud inference services, reducing data-leakage risks and supporting privacy-sensitive research, enterprise deployment, and offline scenarios. Together, hardware adaptability, quantization support, and local agent tooling enable end-to-end agentic workflows within the user's computing environment.

\begin{figure}[t]
    \centering
    \includegraphics[width=\linewidth]{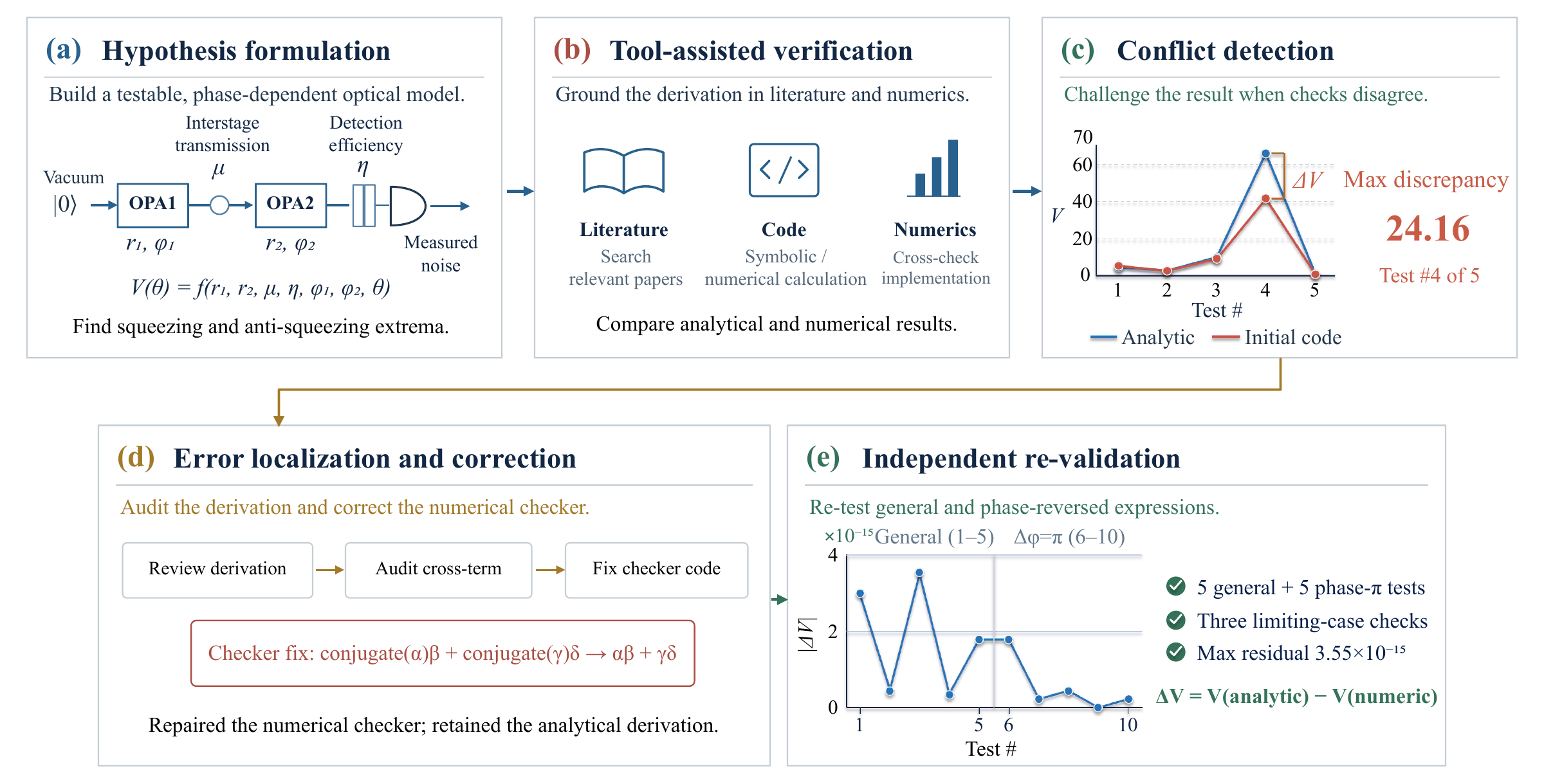}
    \caption{Long-horizon scientific reasoning and self-correction of \ProjectName{} on a quantum optics task from CritPt (\#19). Through hypothesis formulation, tool-assisted verification, conflict detection, error localization and correction, \ProjectName{} identifies and fixes an error in its numerical verification code, resolves discrepancies between its analytical derivation and numerical calculations, and independently re-validates the results.}
    \label{fig:showcase_scientific_research}
\end{figure}

\subsection{Scientific Information Integration and Long-Horizon Reasoning}

\textbf{Deep Research.}
To demonstrate \ProjectName{}'s ability to conduct long-horizon scientific research, we consider an open-ended investigation of the Navier--Stokes problem. The task requires examining its mathematical background, tracing research developments and related claims, and assessing their verification status against academic evidence. Addressing these questions involves extensive information gathering, cross-source comparison, and synthesis across interconnected research topics.

\ProjectName{} organizes this process through a dynamic task graph, decomposing the investigation into dependent research tasks, executing independent inquiries in parallel, and updating the plan as new findings emerge. The results are consolidated through outline planning, section drafting, and iterative revision. In this example, \ProjectName{} produced an eight-section research report with 391 in-text citations, 84 references, and four figures, available in both Chinese and English (Fig.~\ref{fig:showcase_deepresearch}). The Chinese version contains approximately 44,000 characters. This showcase illustrates how \ProjectName{} integrates evidence across multiple research directions into a structured report with traceable sources.

\textbf{Long-Horizon Scientific Reasoning.}
We demonstrate \ProjectName's long-horizon scientific reasoning and self-correction capabilities through a quantum optics problem from CritPt (\#19)~\cite{zhu2026probingcriticalpointcritpt}, as illustrated in Fig.~\ref{fig:showcase_scientific_research}. The task requires deriving the output noise and squeezing extrema of two cascaded optical parametric amplifiers under optical losses. \ProjectName{} first constructs a physical model, consults relevant literature, and verifies its analytical derivation through numerical computation. Upon detecting discrepancies of up to 24.16 between the analytical expression and initial numerical calculations, \ProjectName{} re-examines its derivation and verification code, identifies an incorrect complex conjugation in the correlation cross-term, and corrects the numerical implementation. It then performs ten numerical tests covering both the general expression and the phase-reversed case, reducing the maximum residual to approximately $3.55 \times 10^{-15}$, and further checks consistency with three physical limits. In contrast, Claude Opus 4.8 retains incorrect extremum expressions in the same task. This case highlights \ProjectName's ability to sustain multi-step scientific reasoning, critically examine tool feedback, and iteratively validate its results through a closed loop of \emph{hypothesis formulation, tool-assisted verification, conflict detection, error localization and correction, and independent re-validation}.

\begin{figure}[ht!]
    \centering
    \includegraphics[width=0.85\linewidth,height=0.80\textheight,keepaspectratio]
        {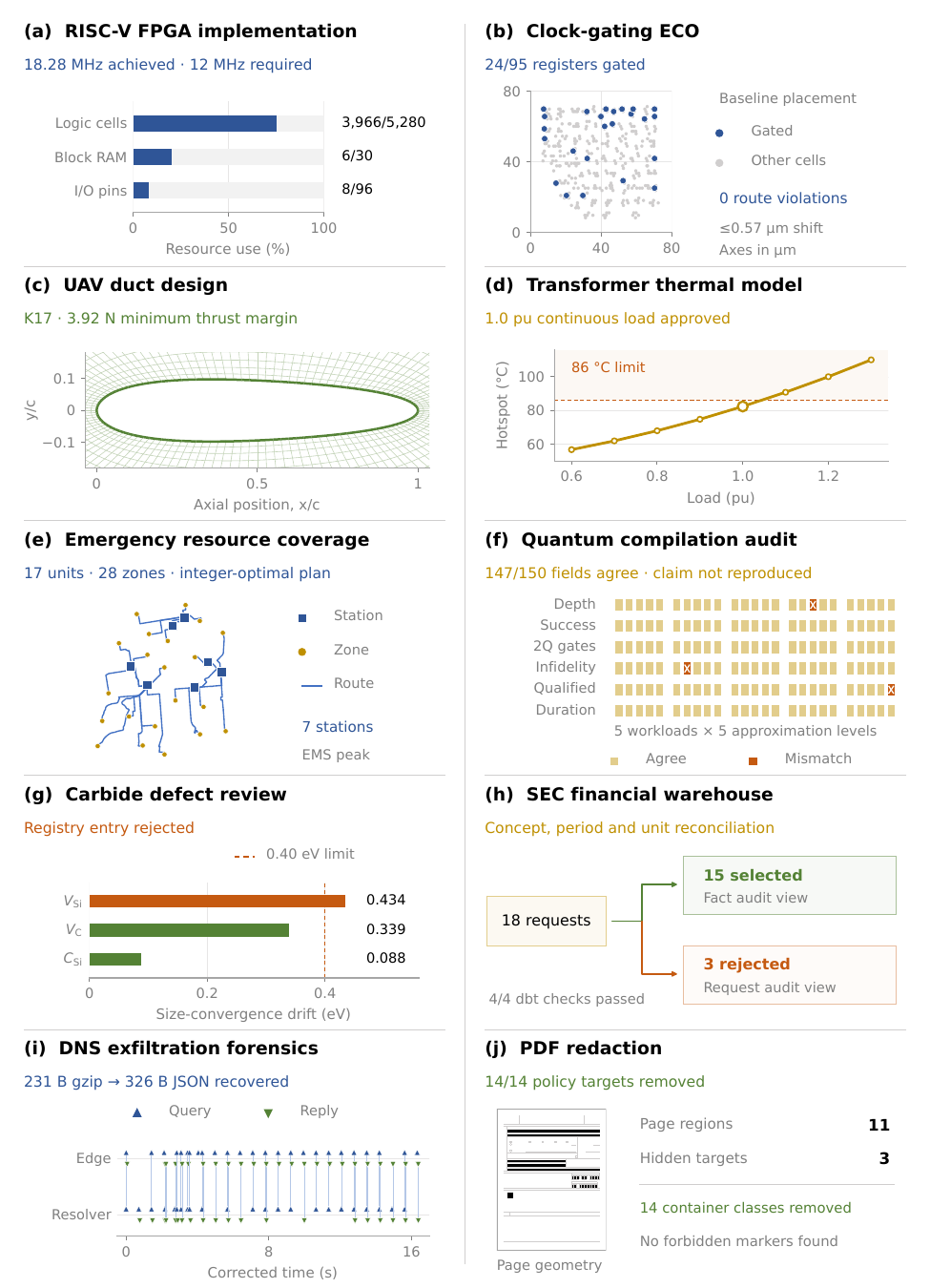}
    \captionsetup{font=small,skip=6pt}
    \caption{
        \textbf{Real-world and professional workflows.}
        Representative artifacts from ten engineering, scientific, planning,
        data and security workflows. Panels show implementation and simulation
        results, coverage plans, audit findings, data reconciliation and
        document processing, with key outcomes highlighted.
    }
    \label{fig:real-world-professional-workflows}
\end{figure}

\subsection{From Scientific Co-working to End-to-end Auto-Research}

\textbf{Real-World and Professional Workflows.}
\textbf{We evaluate \ProjectName{} on ten real-world professional workflows that require domain-specific tools, practical constraints, and verifiable deliverables.} Starting from fixed inputs, the model executes multi-step tasks and produces artifacts that can be independently inspected. These workflows span engineering, scientific computing, planning, financial analysis, cybersecurity, and document processing (Fig.~\ref{fig:real-world-professional-workflows}).

For digital hardware implementation, \ProjectName{} took a PicoRV32-based SoC through logic synthesis, placement and routing, timing closure, and bitstream generation under fixed iCEBreaker board constraints. The design met timing and pin requirements, with configuration checks confirming correct memory initialization. The deliverables included a loadable bitstream, design files, timing reports, and native tool logs.

In the UAV duct study, \ProjectName{} evaluated candidate geometries using SU2, assessing thrust, crossload, power, and pressure recovery across operating conditions. After checking feasibility constraints, it selected K17 and corrected a configuration-serialization precision error by regenerating solver inputs and rerunning the simulations. Other engineering cases included clock-gating ECO implementation, transformer thermal-model calibration, and materials defect registry review.

For emergency-response planning, \ProjectName{} formulated capacity, response-time, coverage, and budget constraints, producing resource-allocation plans with solver certificates and geospatial files. In quantum-compilation auditing, it compared reported results with reproduced workloads, identifying discrepancies that prevented full reproduction of the original claim.

In financial reporting, \ProjectName{} used SEC Companyfacts data to map filing requests to accounting concepts and periods, producing an auditable data warehouse. For network forensics, it reconciled observations from probes with different clock behavior and recovered a DNS transfer payload with supporting packet evidence. In document security, it generated a redacted PDF and verified that the targeted content had been removed.

\begin{figure*}[t]
    \centering
    \includegraphics[width=\textwidth]{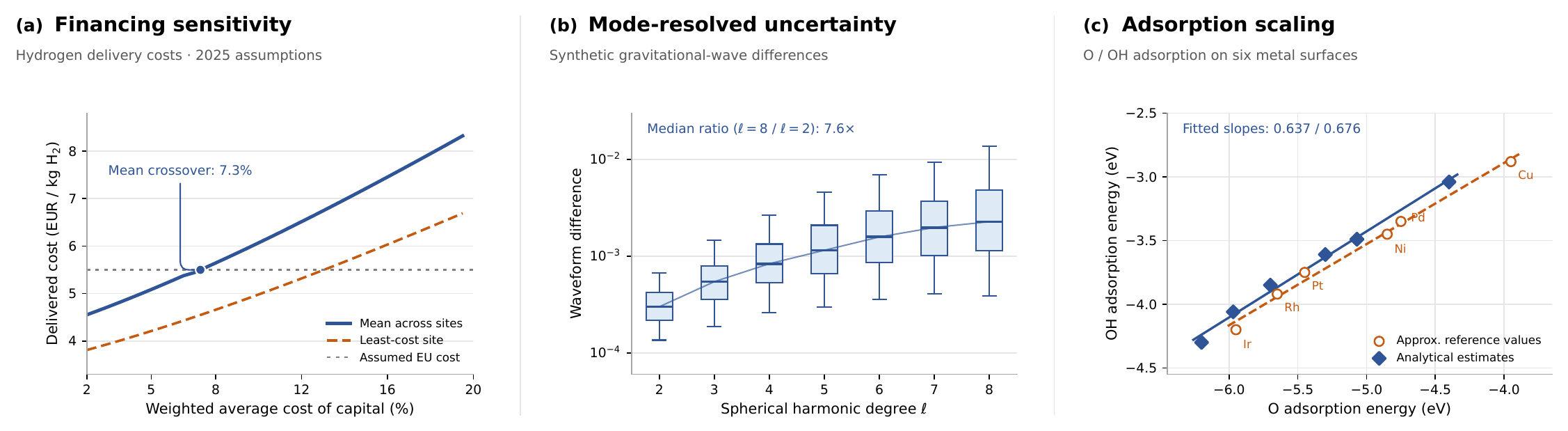}
    \caption{
        \textbf{Computational analyses across three scientific domains.}
        \textbf{(a)} Hydrogen delivery costs across 30 simulated sites under 2025 assumptions. The dashed horizontal line marks the assumed European cost of EUR~5.50/kg.
        \textbf{(b)} Synthetic waveform differences across spherical harmonic modes ($n=1{,}500$ per mode). Boxes show interquartile ranges, center lines indicate medians, and whiskers span the 5th--95th percentiles.
        \textbf{(c)} O/OH adsorption scaling across six metal surfaces using approximate reference values and analytical estimates. Lines show linear fits.
    }
    \label{fig:auto-research-results}
\end{figure*}

\textbf{End-to-end Auto-Research.}
\textbf{\ProjectName~ connected scientific tasks with computational analysis and research reporting across energy economics, gravitational-wave analysis, and atomistic materials.} The resulting workspaces contained analysis code, structured outputs, figures, and reports, enabling inspection of both quantitative results and their underlying assumptions (Fig.~\ref{fig:auto-research-results}). In energy economics, \ProjectName{} modeled green hydrogen delivery from Africa to Europe across 30 simulated production sites, including production, ammonia conversion, shipping, and reconversion. Under the 2025 technology assumptions, saved sensitivity results show that mean delivered cost increases with the weighted average cost of capital. Interpolation of these outputs places the crossover with an assumed European production cost of EUR~5.50 per kilogram at approximately 7.3\% cost of capital. In gravitational-wave analysis, \ProjectName{} characterized resolution, spherical-harmonic-mode, and extrapolation-order errors in supplied synthetic datasets. Mode-resolved distributions show increasing median differences and wider interquartile ranges from $\ell=2$ to $\ell=8$, with a 7.6-fold increase in the median across 1,500 entries per mode. In atomistic materials, \ProjectName{} organized supplied structural parameters and reference results into analyses of water structure, adsorption scaling, and reaction barriers, producing code, visualizations, and a structured research report. For O/OH adsorption on six metal surfaces, linear fits to the approximate reference values and analytically adjusted estimates yielded slopes of 0.637 and 0.676, respectively. The report brought together methodology, quantitative comparisons, and discussion of assumptions and limitations within a single document. These cases illustrate how \ProjectName{} connects domain-specific analyses with organized research outputs, supporting inspection and further development of scientific investigations.

\subsection{Solving Scientific Problems aligned by Scientists}

\begin{figure}[htbp]
    \centering
    \includegraphics[width=0.60\linewidth]{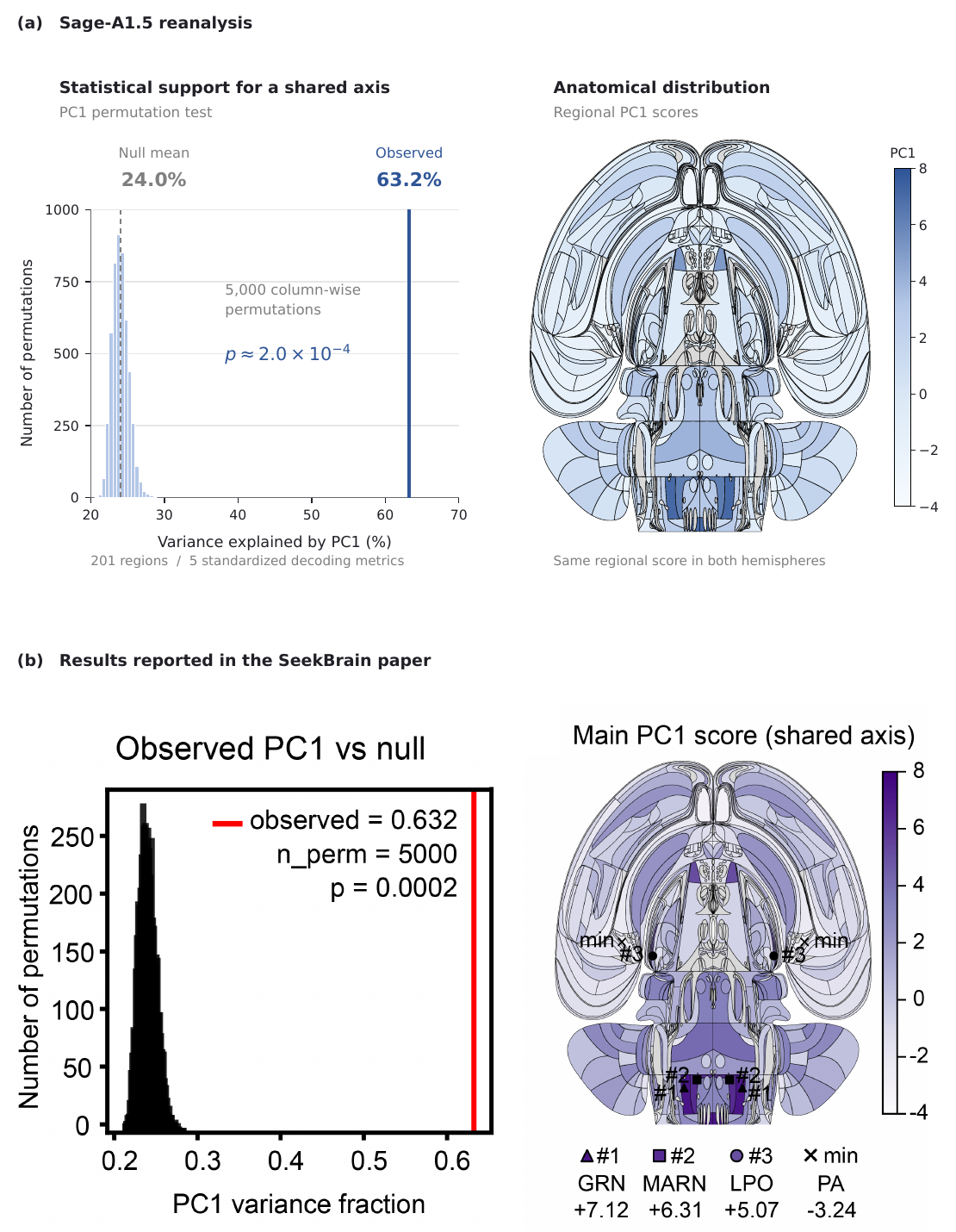}
    \caption{\textbf{Reanalysis of shared decoding structure across mouse brain regions.} \textbf{(a)} Reanalysis by \ProjectName{}. \textbf{(b)} Corresponding results reported in SeekBrain ~\cite{wu2026seekbrain} (Fig.~4c,f). Left: PC1 explains 63.2\% of the variance across five standardized decoding metrics in 201 brain regions, assessed against 5,000 column-wise permutations (empirical $p \approx 2.0 \times 10^{-4}$; null mean in (a), 24.0\%). Right: regional PC1 scores displayed symmetrically on a bilateral Swanson flatmap. Darker blue in (a) and darker purple in (b) indicate higher PC1 scores; regions without a value are gray.}
    \label{fig:neuroscience}
\end{figure}

\begin{figure}[t]
    \centering
    \includegraphics[width=0.9\linewidth]
        {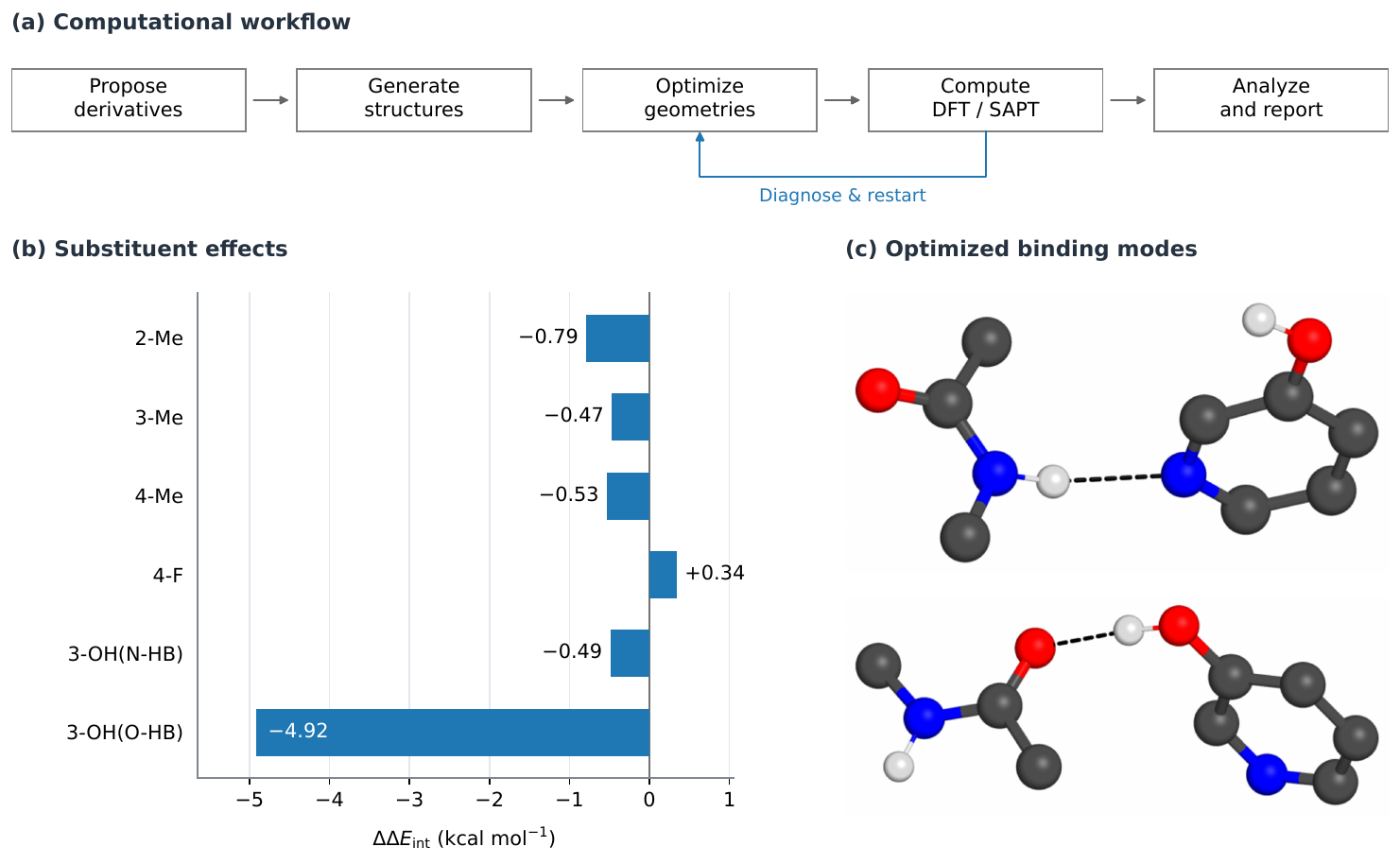}
    \caption{
        \textbf{Substituent effects and binding modes between N-methylacetamide (NMA) and pyridine derivatives in the \ProjectName{} investigation.}
        \textbf{(a)} \ProjectName{} implemented and executed a scientist-guided computational workflow, managing independent jobs concurrently and diagnosing and restarting failed calculations.
        \textbf{(b)} Counterpoise-corrected gas-phase energy differences relative to NMA--pyridine; negative values indicate stronger binding. Each bar shows the lowest-energy sampled pose. The 2-methyl derivative binds more strongly than initially predicted.
        \textbf{(c)} Optimized NMA--3-hydroxypyridine complexes showing ring-nitrogen binding (N-HB, upper) and the more favorable hydroxyl binding (O-HB, lower). Dashed lines indicate hydrogen bonds. Atom colors: C (black), N (blue), O (red), H (white); carbon-bound hydrogens are omitted.
    }
    \label{fig:computational-chemistry}
\end{figure}

\begin{figure}[htbp]
    \centering
    \includegraphics[width=0.85\linewidth]
        {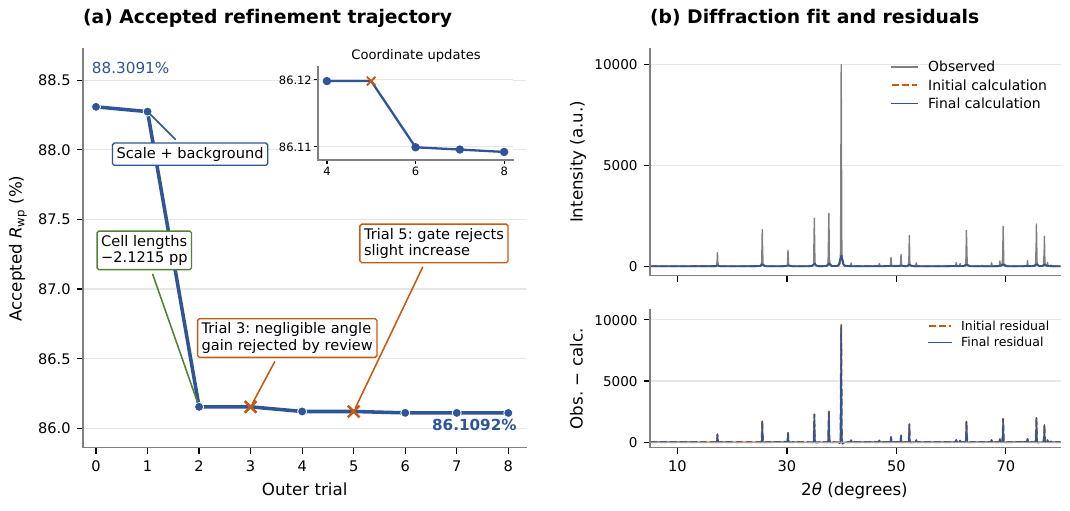}
    \caption{
        \textbf{Feedback-guided refinement of a $\mathrm{Sc_2Au_6}$ structure.}
        \textbf{(a)} Accepted $R_\mathrm{wp}$ over eight trials. Blue circles indicate the initial state and accepted updates, while orange crosses mark rejected trials (3 and 5). The inset highlights late coordinate refinements.
        \textbf{(b)} Observed, initial, and final calculated diffraction profiles, with residuals shown below. Profiles are display-sampled CPU float64 redraws; $R_\mathrm{wp}$ values are from the original NPU run using all 7,500 observations.
    }
    \label{fig:condensed-matter-refinement}
\end{figure}

\textbf{Neuroscience. \ProjectName{} reanalyzed decoding scores from 201 mouse brain regions and reproduced the shared pattern of task and movement decoding reported in the SeekBrain study.} Operating within SeekBrain, a neuroscience-specific multi-agent framework~\cite{wu2026seekbrain}, \ProjectName{} reanalyzed released regional decoding scores from the International Brain Laboratory (IBL) mouse decision-making dataset through a staged workflow encompassing analysis formulation, code execution, validation, and interpretation. This workflow combined principal component analysis (PCA) and movement regression with permutation testing, rank-based robustness checks, and anatomical mapping to the Allen atlas. The first principal component explained 63.2\% of standardized decoding-metric variance, compared with a mean of 24.0\% across 5,000 column-wise permutations (empirical $p \approx 2.0 \times 10^{-4}$). Regional PC1 scores from value-based and rank-based PCA were strongly correlated ($r = 0.960$). Movement-decoding metrics explained approximately 60\% of the across-region variance in choice decoding, while stimulus, choice, and feedback retained distinct residual spatial patterns after accounting for movement. By jointly characterizing shared decoding structure and task-specific residuals, this reanalysis provides a quantitative view of how movement and task information are interwoven across the brain. The close agreement with the findings reported in the SeekBrain paper (Fig.~\ref{fig:neuroscience}) supports \ProjectName{}'s ability to reproduce prior neuroscientific results through agent-driven analysis. This consistency supports the potential to scale the exploration of existing brain-wide datasets and generate testable hypotheses about neural mechanisms.

\textbf{Computational Chemistry. \ProjectName~translated a scientist-specified research plan into an executable quantum chemistry workflow and carried it through calculation, error recovery, analysis, and reporting.}
Given a research question, computational specifications, and a high-level workflow, the \ProjectName-powered agent used Psi4~\cite{smith2020psi4} to investigate how pyridine substitution affects its binding with N-methylacetamide (NMA), a simplified model of a protein--ligand hydrogen-bonding interaction~\cite{spronk2023splinter}. The agent proposed five pyridine derivative candidates, generated structures for 7 isolated monomers and 11 complex poses, and implemented scripts for geometry optimization, binding-energy calculations, and SAPT interaction-energy decompositions~\cite{parker2014sapt}. 

Execution required coordinating dependent computational stages and managing individual jobs over an extended period. The agent scheduled independent calculations concurrently, monitored their progress, inspected failed outputs, and corrected scripts or restarted affected calculations. For example, it recovered a geometry optimization that reached its iteration limit by restarting from saved coordinates with a fresh approximate Hessian. The workflow completed in approximately 12 hours; for context, we estimate that a human researcher implementing the same workflow from scratch would require approximately one working week, including setup, troubleshooting, analysis, and reporting.

After completing the calculations, the agent assembled the results into an illustrated report supported by molecular structures, scripts, and calculation records (Fig.~\ref{fig:computational-chemistry}). Its analysis compared computed outcomes with initial predictions, including stronger-than-expected binding for 2-methylpyridine and distinct binding modes for 3-hydroxypyridine. This case demonstrates sustained execution of a scientist-guided computational investigation, connecting workflow implementation, job management, error recovery, and interpretation within a single research task.

\textbf{Condensed Matter Physics. \ProjectName{} performed autonomous crystal-structure refinement through iterative numerical optimization and feedback-guided parameter updates.}
Refinement from powder X-ray diffraction involves coupled background, lattice, and atomic parameters, requiring careful selection of which variables to update.
Starting from a PXRDgen-generated $\mathrm{Sc_2Au_6}$ structure, \ProjectName{} coordinated three roles: a Planner selected parameter blocks, an Executor ran bounded trials using a constrained Torch-based refinement engine, and a Reviewer assessed the resulting diagnostics. An independent numerical gate checked geometric constraints and the required decrease in the weighted profile residual, $R_\mathrm{wp}$, before committing an update. Each subsequent trial started from the last accepted state (Fig.~\ref{fig:condensed-matter-refinement}).

Across eight trials, the workflow first stabilized scale and background, then tested cell lengths and angles before refining atomic coordinates in blocks. Cell-length refinement produced the largest decrease, 2.1215 percentage points. The third trial's negligible angular improvement was rejected by the Reviewer; in the fifth trial, the numerical gate overrode an acceptance recommendation because $R_\mathrm{wp}$ had slightly increased. Six updates were accepted overall, reducing $R_\mathrm{wp}$ from 88.3091\% to 86.1092\%, an absolute decrease of 2.1999 percentage points. The substantial remaining mismatch limits conclusions about structural accuracy. Nevertheless, this case demonstrates how explicit parameter selection, diagnostic review, and independent acceptance checks can connect sequential refinement decisions to a traceable improvement in the fitting objective.

\begin{figure}[t]
    \centering
    \includegraphics[width=0.80\linewidth]
        {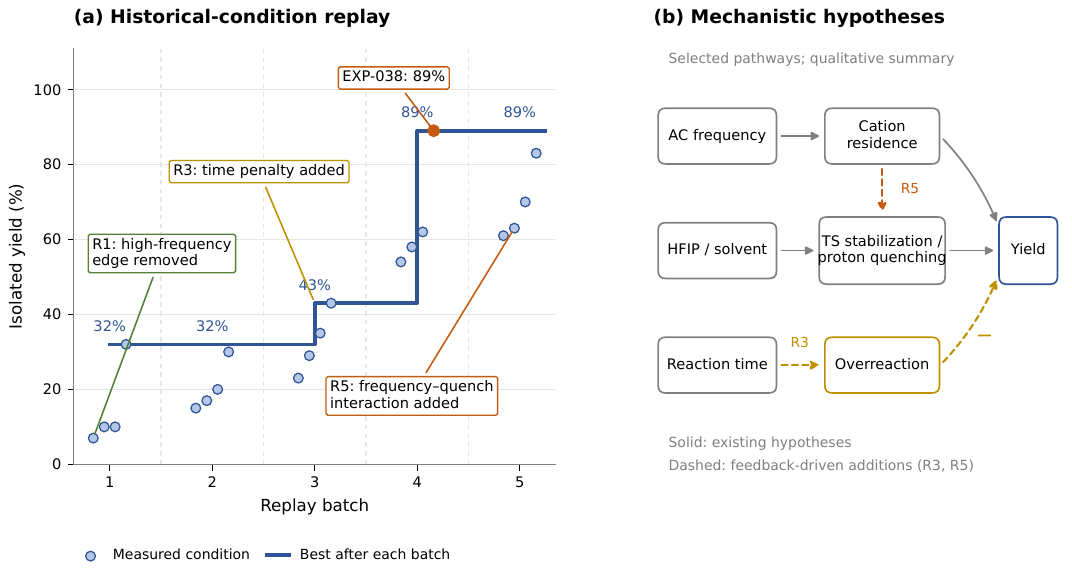}
    \caption{
        \textbf{Mechanistic reasoning during a historical electrochemistry replay.}
        \textbf{(a)} Isolated yields of 20 historical conditions across five batches. The blue staircase tracks the best yield, reaching 89\% at EXP-038 in batch 4. Annotations mark feedback-driven graph updates.
        \textbf{(b)} Evolution of mechanistic hypotheses. Solid arrows show existing connections, while dashed arrows indicate the reaction-time penalty and residence-time--quenching interaction introduced in batches 3 and 5, respectively. TS denotes transition state. These connections represent model hypotheses, not established causal relationships.
    }
    \label{fig:electrochemistry-replay}
\end{figure}
\FloatBarrier


\textbf{Electrochemistry. \ProjectName{} used experimental feedback to revise mechanistic hypotheses and identify a high-yield reaction condition from historical data.}
The case retrospectively examines alternating-current Friedel--Crafts alkylation of 1-bromoethylbenzene using Pt--Pt electrodes, without conducting new experiments. Forty historical records defined the target reaction and an eight-variable search space: solvent, electrolyte, additive, current, frequency, reaction time, arene volume, and electrolyte loading. Categorical choices were restricted to recorded members, and continuous variables to historical ranges. A mechanistic graph informed by 16 literature sources provided candidate-ranking hypotheses. Five batches, each containing four measured conditions, supplied yield feedback for revising those hypotheses (Fig.~\ref{fig:electrochemistry-replay}).

The replay linked changes in observed rankings to specific model edits. Low-frequency observations contradicted an initially favorable high-frequency connection, prompting its removal. Later feedback motivated a reaction-time penalty for overreaction and a proposed interaction between cation residence time and solvent-mediated proton quenching. The cumulative best isolated yield across batches was 32\%, 32\%, 43\%, 89\%, and 89\%. The highest recorded value, EXP-038, used dichloromethane, 1~Hz, 10~mA, 96~min, 500~$\mu$L of arene, and 100~$\mu$L of hexafluoroisopropanol (HFIP). The model associated HFIP with cationic transition-state stabilization and reduced quenching, but the historical comparisons do not establish these effects causally. This case illustrates an interpretable feedback loop in which measured outcomes revise mechanistic working hypotheses while identifying a high-yield condition within the available historical data.

\begin{figure}[t]
    \centering
    \includegraphics[width=\linewidth]
        {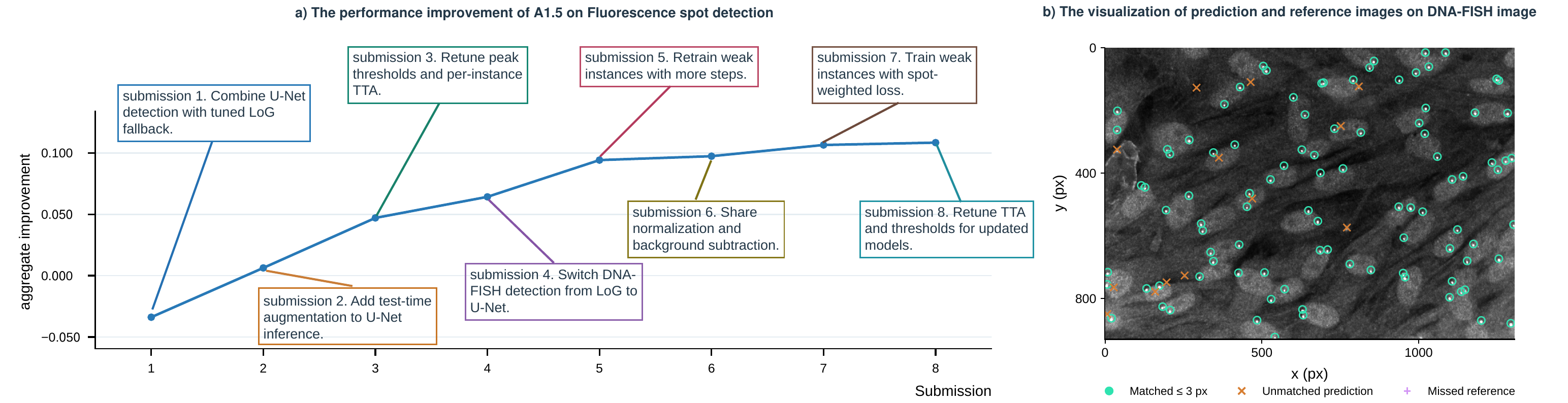}
    \caption{
        \textbf{Iterative optimization of fluorescence spot detection.}
        \textbf{(a)} Performance across eight evaluation submissions, measured by \texttt{aggregate\_improvement} (higher is better). Annotations highlight key implementation changes.
        \textbf{(b)} Predicted and reference spots overlaid on the same DNA-FISH image with identical contrast. Green rings mark matched detections, orange crosses mark unmatched predictions, and purple plus signs mark missed reference spots.
    }
    \label{fig:fluorescence_spot_detection}
\end{figure}

\textbf{Computational Biology. \ProjectName{} iteratively optimized a fluorescence spot detection pipeline, using evaluation feedback to refine model training and inference across diverse microscopy datasets.}
The task covers nine datasets spanning synthetic microscopy, fluorescence in situ hybridization, and live-cell imaging. In the recorded run, \ProjectName{} developed a hybrid pipeline combining U-Net predictions with a Laplacian-of-Gaussian (LoG) fallback. Across eight evaluation submissions, the agent progressively revised its approach, introducing test-time augmentation, adaptive detection thresholds, and changes to the detection pipeline. Later iterations further refined normalization, background subtraction, and training objectives based on evaluation feedback (Fig.~\ref{fig:fluorescence_spot_detection}a). The recorded \texttt{aggregate\_improvement} increased from $-0.0469$ to $0.1158$, reflecting the improvements achieved through successive revisions.

A complementary visualization from a separate retained run compares predicted spots with reference annotations on the same DNA-FISH image (Fig.~\ref{fig:fluorescence_spot_detection}b). The overlay reveals matched detections, unmatched predictions, and missed reference spots, providing a qualitative assessment alongside the benchmark results. This case illustrates how \ProjectName{} uses evaluation feedback to revise an image-analysis pipeline and progressively improve its detection performance.

\section{Limitation and Future Work}

In this report, we have developed Sage-A1.5, a 35B MoE model that shifts the focus of agent training from expanding single-task skills to composing capabilities over long horizons. Rather than relying on fixed rules in an external scaffold, we train the model to organize parallel subagents and verify intermediate evidence within its own execution policy, and then merge specialized domain teachers across search, coding, science, and real-world workflows through unified outcome--process post-training. This provides a practical route for a compact open model to tackle complex, multi-stage engineering and scientific problems.

During this effort, we also observed that organizing long-horizon work requires more than task delegation and mid-course verification. On the execution side, our current orchestration and verification training still centers on search and research synthesis, while ultra-long trajectories can still accumulate heavy context overhead; meanwhile, the unified model does not yet fully retain the peak performance of every domain teacher after joint post-training. On the application side, our scientific showcases still operate within computational simulations and historical data replays rather than active physical experiments. In future work, we will use intermediate verification points as natural boundaries to compress completed exploration history into concise stage records, extend learned orchestration to broader engineering workflows while reducing cross-domain distillation loss, and work with scientists to connect Sage-A1.5 with real-world experimental validation.


\clearpage
\begingroup
\sloppy
\bibliographystyle{unsrtnat}
\bibliography{main}
\endgroup

\clearpage
\appendix
\section{Appendix}

\subsection{Contributions and Acknowledgments}

\definecolor{damaiblue}{RGB}{10, 102, 155}
\definecolor{damaiorange}{RGB}{180,50,50}
\definecolor{damaired}{RGB}{10, 50, 50}

Within each block, sort names in reverse alphabetical order (Z–A) by first name: \\

\noindent
\textbf{\color{damaired} Lead Authors} \\ [2.5mm]
Wenjie Lou, Tianshuo Peng, Shiyang Feng, Shengji Tang, Runmin Ma, Ruizhe Chen, Jinxin Shi, Haoyang Peng, Fangchen Yu

\noindent
\textbf{\color{damaired} Core Authors} \\ [2.5mm]
Zhijie Zhong, Yuhao Zhou, Yue Fan, Yixin Chen, Yiming Wu, Yiheng Wang, Yigeng Jiang, Yichen Jiang, Yi Xie, Xu Guo, Xiaohan He, Tao Hu, Songtao Huang, Siyu Zhou, Shuaiyu Zhang, Shangheng Du, Qinglong Cao, Kun Li, Jingqi Ye, Hongbin Zhou, Guohang Yan

\noindent
\textbf{\color{damaired} Contributors} \\ [2.5mm]
Zongsheng Cao, Ziyi Wang, Zhuo Liu, Zhiyin Yu, Zhiwang Zhou, Zelin Tan, Yuqing Zhu, Yaoze Zhang, Yang Chen, Xuming He, Xiuliang Liang, Xingjian Guo, Xiang Zhuang, Weiping Shen, Weida Wang, Tianyuan Zhou, Tianlin Ye, Tengyuan Fan, Siyu Yan, Shuo Li, Shuchen Pu, Shengyuan Xu, Shao Zhang, Ran Wei, Qinyuan Qin, Qingyu Yang, Nianchen Deng, Meng Dai, Kuikun Liu, Kaitao Chen, Juncheng Xiao, Jiangning Liu, Jianbiao Mei, Jia Yu, Haonan He, Hangfan Zhang, Cheng Yang, Chen Zhang, Bihao Zhan, Benteng Cheng

\noindent
\textbf{\color{damaired} Scientific Directors} \\  [2.5mm]
Yuqiang Li, Yihao Liu, Yang Yang, Xujun Fu, Xiaosong Wang, Xiangyu Yue, Wenwei Zhang, Shuyue Hu, Pinlong Cai, Mianxin Liu, Lei Bai, Jie Zhou, Jie Hou, Jiamin Wu, Hui Wang, Haijun Li, Guangyu Wang, Dongzhan Zhou, Chunfeng Song, Chao Yang, Biqing Qi

\noindent
\textbf{\color{damaired} Advisors } \\  [2.5mm]
Xing Xie, Dahua Lin, Bowen Zhou

\noindent
\textbf{\color{damaired} Project Co-lead} \\  [2.5mm]
Xiangchao Yan, Wenlong Zhang, Shufei Zhang, Peng Ye, Bo Zhang \\

\end{document}